\documentclass[sigconf]{acmart}
\usepackage{enumitem}
\usepackage{multirow,multicol}
\usepackage{booktabs}
\usepackage{colortbl}
\AtBeginDocument{%
  }

\copyrightyear{2026}
\acmYear{2026}
\setcopyright{cc}
\setcctype{by}
\acmConference[KDD '26]{Proceedings of the 32nd ACM SIGKDD Conference on Knowledge Discovery and Data Mining V.2}{August 09--13, 2026}{Jeju Island, Republic of Korea}
\acmBooktitle{Proceedings of the 32nd ACM SIGKDD Conference on Knowledge Discovery and Data Mining V.2 (KDD '26), August 09--13, 2026, Jeju Island, Republic of Korea}
\acmDOI{10.1145/3770855.3818948}
\acmISBN{979-8-4007-2259-2/2026/08}

\begin{document}

\title{BatteryMFormer: Multi-level Learning for Battery Degradation Trajectory Forecasting}

\author{Ruifeng Tan}
\authornote{Both authors contributed equally to this research.}
\affiliation{%
  \institution{Sustainable Energy and Environment Thrust, The Hong Kong University of Science
		and Technology (Guangzhou)}
  \city{Guangzhou}
  \country{China}
}
\email{rtan474@connect.hkust-gz.edu.cn}

\author{Jintao Dong}
\authornotemark[1]
\affiliation{%
  \institution{School of Computer Science and Engineering, Central South University}
  \city{Changsha}
  \country{China}}
\email{jintaodong@csu.edu.cn}

\author{Weixiang Hong}
\affiliation{%
  \institution{Sustainable Energy and Environment Thrust, The Hong Kong University of Science
		and Technology (Guangzhou)}
  \city{Guangzhou}
  \country{China}
}
\email{whong719@connect.hkust-gz.edu.cn}

\author{Jia Li}
\authornote{Corresponding authors.}
\affiliation{%
	\institution{Data Science and Analytics Thrust, The Hong Kong University of Science
		and Technology (Guangzhou)}
	\city{Guangzhou}
	\country{China}
}
\email{jialee@ust.hk}

\author{Jiaqiang Huang}
\authornotemark[2]
\affiliation{%
  \institution{Sustainable Energy and Environment Thrust, The Hong Kong University of Science
		and Technology (Guangzhou)}
  \city{Guangzhou}
  \country{China}
}
\email{seejhuang@hkust-gz.edu.cn}

\author{Tong-Yi Zhang}
\authornotemark[2]
\affiliation{%
  \institution{Material Genome Institute, \\Shanghai University}
  \city{Shanghai}
  \country{China}
}
\affiliation{%
  \institution{Advanced Materials Thrust and Sustainable Energy and Environment Thrust, The Hong Kong University of Science and Technology (Guangzhou)}
  \city{Guangzhou}
  \country{China}
}
\email{mezhangt@hkust-gz.edu.cn}
\renewcommand{\shortauthors}{Ruifeng Tan et al.}

\begin{abstract}
Early battery degradation trajectory forecasting (BDTF), which predicts the full-life state-of-health trajectory from early operational data, is critical for battery optimization, manufacturing, and deployment. Battery degradation data exhibit two key characteristics. First, degradation data present a multi-level structure, including regularities shared within aging conditions and trajectory patterns shared across batteries. Second, degradation-related variations in voltage-current profiles are often localized to specific state of charge (SOC) intervals. Existing approaches often fail to explicitly model these characteristics. To bridge this gap, we propose BatteryMFormer, a multi-level Transformer for early BDTF. BatteryMFormer integrates (1) an aging-condition-aware decoder that injects aging-condition priors via aging-condition-informed queries and aging-condition-aware attention, (2) a meta degradation pattern memory that learns and retrieves trajectory prototypes to guide long-horizon forecasting, and (3) a dual-view encoder that jointly captures temporal dynamics and SOC-localized variations from voltage and current time series. Extensive experiments on four battery domains show that BatteryMFormer consistently outperforms state-of-the-art baselines, marking a significant step toward reliable BDTF. Our code is available at \url{https://github.com/Ruifeng-Tan/BatteryMFormer}.
\end{abstract}

\ccsdesc[500]{Information systems~Data mining}

\keywords{materials informatics, battery informatics, time series}


\maketitle

\section{Introduction}
Rechargeable batteries are ubiquitous in modern industry, powering applications ranging from electric vehicles and grid-scale energy storage to portable electronics \cite{20221311845974,tao2023collaborative,zhang2025unlocking,tan2025PBT}. In 2024, global battery shipments exceeded 1545~GWh and are projected to reach 4700~GWh by 2030 \cite{ZHENG2026114180,fleischmann2023battery}. This rapid expansion highlights the need for advanced modeling frameworks to support battery optimization, manufacturing, and deployment \cite{LiPM,TAN2024103725,zhang2025battery,20191406719359,attia2020closed}. In particular, battery degradation trajectory forecasting (BDTF), which predicts battery state-of-health (SOH) trajectories from beginning of life to end of life, occupies a critical frontier. By forecasting full-life degradation trajectory from early-stage operational data, BDTF enables accelerated degradation assessment and timely maintenance for battery-powered systems \cite{TAN2024103725,LI2021230024,HUANG2026239148}.

\begin{figure}[t]
    \centering
    \includegraphics[width=1.0\linewidth]{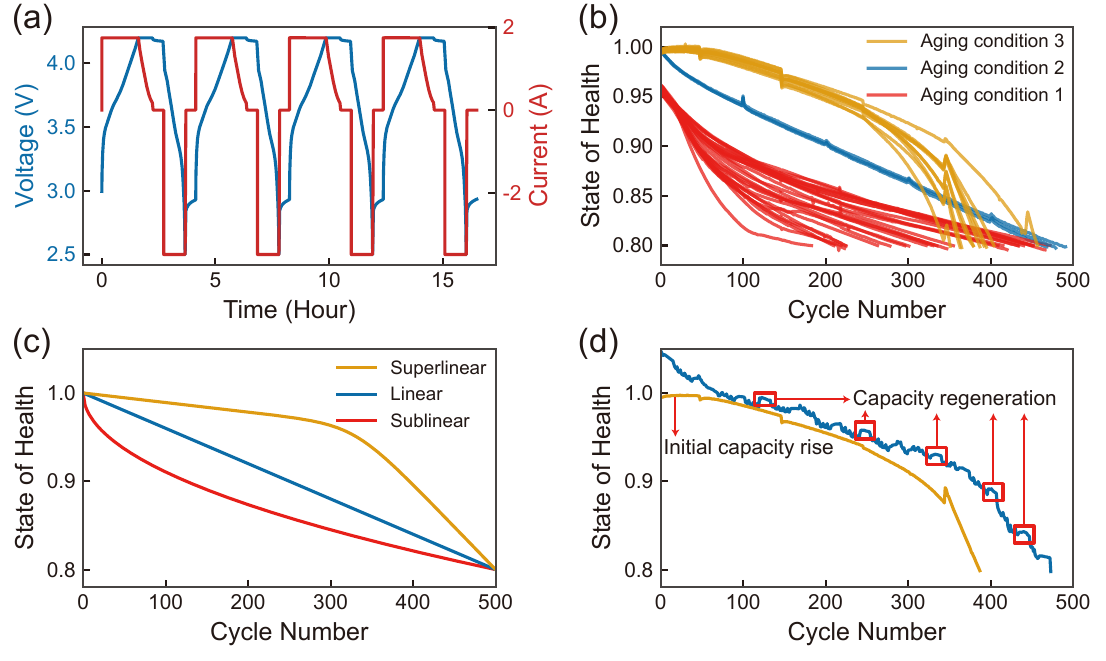}
    \caption{\textbf{Motivation for multi-level learning in BDTF.} (a) An example of partial operational voltage and current time series. (b) SOH trajectories under different aging conditions. (c) Schematic of three canonical trajectory shapes. (d) Examples of additional trajectory phenomena beyond the canonical shapes.}
    \label{Fig1}
\end{figure}

Machine learning (ML) models have recently emerged as promising solutions to BDTF. Existing approaches primarily fall into feature-engineering-based methods and representation-learning-based methods. Feature-engineering-based methods typically extract handcrafted descriptors from voltage and current time series (Figure \ref{Fig1}a) \cite{20250417745793,10332202,10491306}, whereas these features are often protocol-specific or dataset-specific and may be unavailable or ineffective across diverse aging conditions. Representation-learning-based methods instead focus on learning mappings from raw measurements to future SOH trajectories \cite{LI2021230024,LI2022453,TAN2024103725,LIU2025114736,HUANG2026239148,BatteryLife,HUANG2024122825,20254519447317}. An intuitive modeling choice is to treat BDTF as generic time-series forecasting and extrapolate future SOH from historical SOH using generic time series forecasters (e.g. Informer \cite{Informer}) \cite{LI2021230024,LI2022453,TAN2024103725,LIU2025114736,20254519447317}. While effective in some settings, early-cycle SOH can be nearly indistinguishable across batteries whose long-horizon trajectories diverge substantially, and therefore forecasting with SOH as the only input can be unsuitable for early BDTF (Figure \ref{Fig1}b). This limitation has motivated growing interest in models that exploit fine-grained voltage–current profiles for forecasting \cite{HUANG2026239148,BatteryLife,HUANG2024122825}.

Despite these advances, current models still exhibit two critical research gaps.
First, these methods operate at the \emph{battery level} and do not explicitly model the \emph{multi-level structure} of degradation.
Batteries under the same aging condition (e.g., specifications, formation, and operating conditions) exhibit consistent operational patterns, and prior work shows that batteries under similar aging conditions can be characterized by a small set of handcrafted descriptors \cite{20191406719359,20214611166935,20232714333821,20250417745793,LI2022453}.
However, existing models fail to promote \emph{aging-condition-consistent} representations.
Moreover, although trajectories appear diverse, established battery knowledge \cite{attia2022knees} suggests that their \emph{global shapes are highly structured} and often fall into a small family of patterns linked to common mechanisms (Figure \ref{Fig1}c).
Additional phenomena such as initial capacity rise \cite{20191406719359} and capacity regeneration \cite{HUANG2024122825} can occur (Figure \ref{Fig1}d), but the space of plausible trajectories remains constrained.
Second, degradation-relevant variations in voltage–current profiles often concentrate within specific SOC intervals (Figure \ref{Fig2}), as underlying electrochemical mechanisms can be manifested as localized electrochemical signal variations along the SOC axis (e.g., phase transition) \cite{TAN2024103725,BIRKL2017373}.
Nevertheless, most methods either emphasize temporal modeling or treat SOC intervals uniformly, diluting localized signals.

To address these limitations, we propose \textbf{BatteryMFormer} (Battery Multi-level Transformer), a novel deep learning architecture that integrates multi-level learning across aging conditions, trajectory patterns, and battery-specific representations. BatteryMFormer consists of three major components: (1) Aging-condition-aware decoder that injects aging-condition priors via aging-condition-informed queries and aging-condition-aware attention to promote aging-condition-consistent representations; (2) Meta degradation pattern memory that learns and retrieves prototypical trajectory patterns to guide long-horizon forecasting; and (3) Dual-view encoder that captures complementary temporal dynamics and SOC-localized variations from voltage-current profiles.

The main contributions of this paper are summarized as follows:
\begin{itemize}[left=0pt]
    \item We identify and formalize the multi-level structure of early BDTF, including aging-condition regularities, trajectory patterns shared across batteries, and SOC-localized degradation signatures in operational data.
    \item We propose BatteryMFormer, a multi-level Transformer that integrates (i) an aging-condition-aware decoder, (ii) a meta degradation pattern memory, and (iii) a dual-view encoder with temporal and SOC perspectives.
    \item We conduct extensive experimental evaluation, the results from which demonstrate the superior performance of our approach across four battery domains from the largest public real-world battery lifetime database.
\end{itemize}

\section{Preliminaries}
\subsection{Aging Condition}\label{SecAging}
\begin{figure}[!t]
    \centering
    \includegraphics[width=1.0\linewidth]{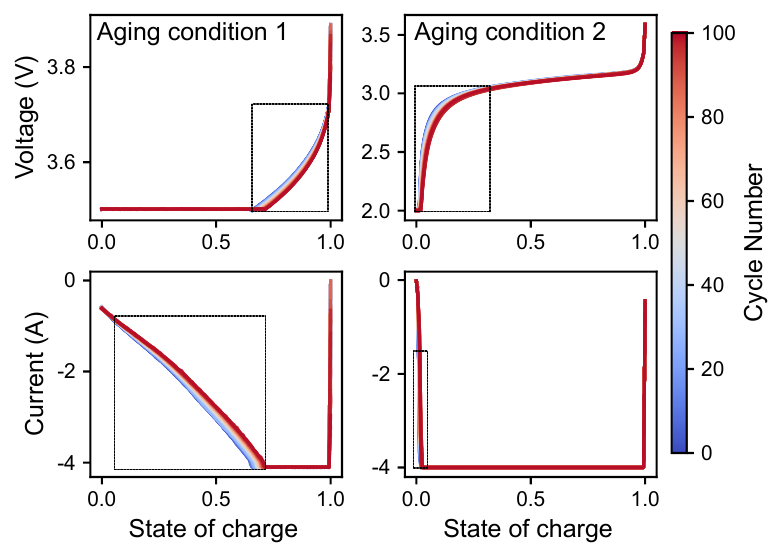}
    \caption{\textbf{SOC-localized degradation signatures in voltage--current profiles.} Voltage--SOC (top) and current--SOC (bottom) curves over the first 100 cycles for two representative aging conditions. Although the global profiles evolve smoothly with cycling, aging-induced deviations can concentrate within specific SOC intervals (dashed boxes).}
    \label{Fig2}
\end{figure}

We use aging condition to denote the recorded experimental settings and battery specifications that determine a battery’s degradation regime. In this work, an aging condition is represented as a tuple of aging factors, including positive electrode, negative electrode, electrolyte, package structure, nominal capacity, manufacturer, formation protocol, charge protocol, discharge protocol, and operating temperature. Different factor tuples correspond to different aging conditions. Batteries operated under different aging conditions can exhibit distinct degradation trajectories (Figure \ref{Fig1}b) and patterns of voltage--current profiles (Figure \ref{Fig2})~\cite{BatteryLife,zhang2025battery,tan2025PBT}.

\subsection{Degradation Trajectory}\label{SecDegradation}
Degradation trajectories are measured from repeated cycles, with each having a charge and discharge process. Following prior work \cite{BatteryLife,20223412623287,20191406719359}, we compute the discharge capacity of cycle $i$ as
\begin{equation}
Cap_i=\int_{t_1}^{t_2}|I(t)|\,dt,
\label{EqCapacity}
\end{equation}
where $t_1$ and $t_2$ denote the start and end times of the discharge process, and $I(t)$ is the measured current at time $t$, with $|I(t)|$ used to make the definition invariant to sign conventions. The state of health (SOH) at cycle $i$ is defined as
\begin{equation}
\mathrm{SOH}_i=\frac{Cap_i}{Cap_0\times DoD},
\end{equation}
where $DoD$ is the depth of discharge, and $Cap_0$ denotes the nominal capacity for all datasets except CALB, where $Cap_0$ is defined as the first-cycle discharge capacity following the CALB protocol in BatteryLife \cite{BatteryLife}.

\subsection{Task Formulation}\label{SecTaskFormulation}
Following prior work~\cite{zhang2025battery,20191406719359,BatteryLife}, we use the first $S\le 100$ cycles as the early stage and forecast the SOH trajectory beyond the observation window.
We denote by $\boldsymbol{a}$ the available aging-condition metadata of a battery, including recorded experimental settings and specifications.
Let $\mathbf{X}_i$ denote the cycle-$i$ operational data, consisting of voltage and current time series (and any auxiliary variables derived from $\boldsymbol{a}$ and these early-cycle measurements, e.g., capacity and SOC).
We define the early input as ordered sequences
\begin{equation}
\mathbf{G}_{1:S}=\bigl(\mathbf{X}_{1:S},\,\boldsymbol{a}\bigr), 
\qquad 
\mathbf{X}_{1:S}=[\mathbf{X}_1,\ldots,\mathbf{X}_S].
\end{equation}
We use $t_{\mathrm{eol}}$ to denote the end-of-life (EOL) cycle index, defined as the first cycle at which $\mathrm{SOH}$ falls below a threshold $\tau$ (Appendix~\ref{AppendixEOL}).
Let $\mathbf{y}_{1:t_{\mathrm{eol}}}\in\mathbb{R}^{t_{\mathrm{eol}}}$ denote the measured SOH trajectory.
The goal of early BDTF is to learn a forecasting model $f(\cdot)$ that predicts the future SOH trajectory given the first $S$ cycles:
\begin{equation}
\hat{\mathbf{y}}_{S+1:t_{\mathrm{eol}}} = f(\mathbf{G}_{1:S}).
\end{equation}

\begin{figure*}
    \centering
    \includegraphics[width=0.85\linewidth]{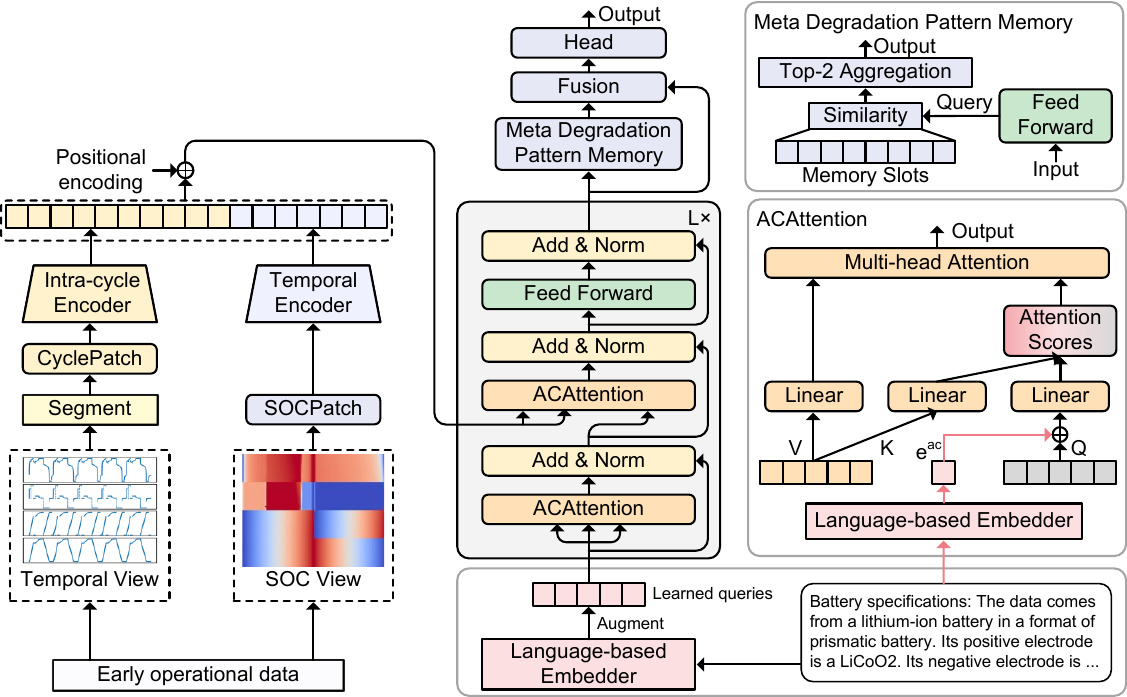}
    \caption{\textbf{An overview of BatteryMFormer.} Left: dual-view encoder (temporal and SOC views). Middle: aging-condition-aware decoder. Right: details of meta degradation pattern memory and aging-condition-aware attention.}
    \label{Figframework}
\end{figure*}
\section{Methodology}
Figure \ref{Figframework} presents the overall architecture of BatteryMFormer, a Transformer with multi-level inductive biases for early BDTF.
BatteryMFormer encodes early operational data into complementary temporal and SOC tokens via a dual-view encoder (Section \ref{SecDualView}), refines these tokens with an aging-condition-aware decoder (Section \ref{SecAWTrans}), and retrieves prototypical trajectory patterns from a meta degradation pattern memory (Section \ref{SecMemory}) to guide long-horizon forecasting.

\subsection{Dual-View Encoder}\label{SecDualView}
The dual-view encoder maps early operational data into temporal-view and SOC-view tokens.
Following BatteryLife~\cite{BatteryLife}, we obtain within-cycle capacity via ampere-hour counting from current time series to encode temporal information and additionally compute SOC (Appendix~\ref{AppendixSOC}).
After resampling each cycle to $L$ data points, the first $S$ cycles are represented as $\mathbf{X}\in\mathbb{R}^{S\times L\times 4}$ with variables (voltage, current, capacity, SOC).

\textbf{SOC view.}
To capture SOC-localized degradation signatures (Figure~\ref{Fig2}), we construct SOC-view tokens by modeling cross-cycle evolution within each SOC interval.
Given $\mathbf{X}\in\mathbb{R}^{S\times L\times 4}$, where each cycle contains $L$ SOC-aligned points and 4 variables, we reshape the $i$-th cycle as $\mathbf{X}_i\in\mathbb{R}^{4\times L}$, treating variables as channels.
We then apply a 1D convolution along the SOC axis:
\begin{equation}
\hat{\mathbf{Z}}_i=\mathrm{Conv1D}(\mathbf{X}_i)\in\mathbb{R}^{d\times M},\qquad
M=\left\lfloor\frac{L-P}{P}\right\rfloor+1,
\end{equation}
where $P$ is both the patch length and stride.
Stacking all cycles yields $\hat{\mathbf{Z}}\in\mathbb{R}^{S\times d\times M}$.
For each SOC interval $m$, we collect $\hat{\mathbf{Z}}_{:,:,m}\in\mathbb{R}^{S\times d}$ across cycles and feed it into a shared temporal encoder implemented with feed-forward neural networks and GELU activations~\cite{hendrycks2016gaussian}.
The encoder aggregates information along the cycle axis and produces one SOC token:
\begin{equation}
\mathbf{t}^{\mathrm{soc}}_m=\mathrm{TempEnc}(\hat{\mathbf{Z}}_{:,:,m})\in\mathbb{R}^{d}.
\end{equation}
Concatenating all interval tokens yields $\mathbf{T}^{\mathrm{soc}}=[\mathbf{t}_1^{\mathrm{soc}};\ldots;\mathbf{t}_M^{\mathrm{soc}}]\in\mathbb{R}^{M\times d}$.

\textbf{Temporal view.}
In parallel to the SOC view, we construct a temporal view that summarizes each early cycle as a cycle-level token to capture intra-cycle dynamics.
Following CyclePatch~\cite{BatteryLife}, we project the resampled multivariate series of cycle $i$ into a $d$-dimensional embedding and refine it with an intra-cycle encoder:
\begin{gather}
\hat{\mathbf{X}}_i = \mathrm{CyclePatch}(\mathbf{X}_i)
= \mathrm{Flatten}(\mathbf{X}_i)\mathbf{W}+\mathbf{b}, \\
\mathbf{z}^{\mathrm{temporal}}_{i} = \mathrm{Intra\text{-}CycleEncoder}(\hat{\mathbf{X}}_i), \qquad i=1,\ldots,S,
\end{gather}
where $\mathbf{W}$ and $\mathbf{b}$ are learnable parameters.
Stacking $\{\mathbf{z}^{\mathrm{temporal}}_{i}\}_{i=1}^{S}$ yields temporal tokens
\begin{equation}
\mathbf{H}^{\mathrm{temporal}}
=
[\mathbf{z}^{\mathrm{temporal}}_{1};\ldots;\mathbf{z}^{\mathrm{temporal}}_{S}]
\in\mathbb{R}^{S\times d}.
\end{equation}
We further inject cycle-level descriptors by projecting $\mathbf{X}_f\in\mathbb{R}^{S\times d_f}$ to the token space and adding it to $\mathbf{H}^{\mathrm{temporal}}$:
\begin{equation}
\mathbf{T}^{\mathrm{temporal}}=\mathbf{H}^{\mathrm{temporal}}+\mathbf{X}_{f}\mathbf{W}_f+\mathbf{b}_f,
\end{equation}
where $\mathbf{W}_f$ and $\mathbf{b}_f$ are learnable parameters.
In this work, $\mathbf{X}_f$ consists of Coulombic efficiency and energy efficiency, which are commonly available on a per-cycle basis.

Together, $\mathbf{T}^{\mathrm{temporal}}\in\mathbb{R}^{S\times d}$ and $\mathbf{T}^{\mathrm{soc}}\in\mathbb{R}^{M\times d}$ provide complementary inputs for subsequent decoding.

\subsection{Aging-Condition-Aware Decoder}\label{SecAWTrans}
Batteries operated under the same/similar aging conditions often exhibit consistent/similar degradation signatures~\cite{20191406719359,20214611166935,20232714333821,20250417745793}.
To exploit such aging-condition-level regularities, we design an aging-condition-aware decoder (ACDecoder) with two mechanisms:
(i) \emph{aging-condition-informed queries}, which inject an aging-condition prior into the decoder states, and
(ii) \emph{aging-condition-aware attention}, which conditions attention on the aging-condition prior.

\textbf{Aging-condition-informed queries.}
Inspired by~\cite{MedSpaformer}, ACDecoder starts from learnable generic queries $\mathbf{Q}_g\in\mathbb{R}^{\bar{s}\times d}$ and injects aging-condition information via additive conditioning.
Let $\boldsymbol{a}$ denote the structured aging-condition metadata (Section~\ref{SecAging}) and let $\pi(\boldsymbol{a})$ be the corresponding metadata-to-text prompt~\cite{tan2025PBT}.
We encode $\pi(\boldsymbol{a})$ using a language-based embedder:
\begin{gather}
\mathbf{z}^{ac}=\mathrm{LastValid}\bigl(\mathrm{Enc}(\pi(\boldsymbol{a}))\bigr)\in\mathbb{R}^{d_{\mathrm{enc}}}, \\
\mathbf{e}^{ac}=\mathbf{z}^{ac}\mathbf{W}_1+\mathbf{b}_1\in\mathbb{R}^{d},
\end{gather}
where $\mathrm{Enc}(\cdot)$ is a language-based embedder (Qwen3-Embedding-0.6B~\cite{qwen3embedding}), $\mathrm{LastValid}(\cdot)$ retrieves the embedding of the last non-padding token, and $\mathbf{W}_1\in\mathbb{R}^{d_{\mathrm{enc}}\times d}$ and $\mathbf{b}_1\in\mathbb{R}^{d}$ are learnable parameters.
We then project $\mathbf{e}^{ac}$ to produce one prior vector per query token:
\begin{gather}
\hat{\mathbf{e}}^{ac}_i=\mathbf{e}^{ac}\mathbf{W}_{2,i}+\mathbf{b}_{2,i}\in\mathbb{R}^{d},\qquad i=1,\ldots,\bar{s}, \\
\hat{\mathbf{E}}^{ac}=[\hat{\mathbf{e}}^{ac}_1;\ldots;\hat{\mathbf{e}}^{ac}_{\bar{s}}]\in\mathbb{R}^{\bar{s}\times d}, \\
\mathbf{X}_{de}=\mathbf{Q}_g+\hat{\mathbf{E}}^{ac}.
\label{eq:ac_queries}
\end{gather}
Here $\mathbf{W}_{2,i}\in\mathbb{R}^{d\times d}$ and $\mathbf{b}_{2,i}\in\mathbb{R}^{d}$ are learnable parameters.
Each $\hat{\mathbf{e}}^{ac}_i$ provides a query-specific prior, yielding aging-condition-informed queries (ACQuery) $\mathbf{X}_{de}$ for conditioning different queries on different aspects of the aging-condition information.

\textbf{Aging-condition-aware attention.}
Beyond query initialization, ACDecoder promotes aging-condition-consistent attention by modulating queries with $\hat{\mathbf{E}}^{ac}$.
Given queries $\mathbf{Q}$ and key--value tokens $(\mathbf{K},\mathbf{V})$, we define aging-condition-aware attention (ACAttention) as follows:
\begin{gather}
\mathrm{ACAttention}(\mathbf{Q},\mathbf{K},\mathbf{V},\hat{\mathbf{E}}^{ac})
=
\mathrm{Concat}(\mathrm{head}_1,\ldots,\mathrm{head}_h)\mathbf{W}^{O},\\
\mathrm{head}_i
=
\mathrm{Attention}\bigl((\mathbf{Q}+\hat{\mathbf{E}}^{ac})\mathbf{W}_i^{Q},\ \mathbf{K}\mathbf{W}_i^{K},\ \mathbf{V}\mathbf{W}_i^{V}\bigr).
\end{gather}
Here $\mathrm{Attention}(\cdot)$ is the attention in standard Transformer~\cite{20182105216120}. This query modulation injects aging-condition priors into every attention operation, thereby promoting aging-condition-consistent decoding throughout the network.

\textbf{ACDecoder layer.}
Let $\mathbf{T}^{\mathrm{temporal}}\in\mathbb{R}^{S\times d}$ and $\mathbf{T}^{\mathrm{soc}}\in\mathbb{R}^{M\times d}$ be the dual-view tokens (Section~\ref{SecDualView}), and $\mathbf{T}=[\mathbf{T}^{\mathrm{temporal}};\mathbf{T}^{\mathrm{soc}}]+\mathbf{P}\in\mathbb{R}^{(S+M)\times d}$, where $\mathbf{P}\in\mathbb{R}^{(S+M)\times d}$ is positional encoding \cite{20182105216120}.
With $\mathbf{H}^{0}=\mathbf{X}_{de}\in\mathbb{R}^{\bar{s}\times d}$, the process in the $l$-th ACDecoder layer is
\begin{gather}
\mathbf{H}^{l}_{1}
=
\mathrm{LN}\Bigl(
\mathbf{H}^{l-1}
+
\mathrm{ACAttention}\bigl(
\mathbf{H}^{l-1},\mathbf{H}^{l-1},\mathbf{H}^{l-1},\hat{\mathbf{E}}^{ac}
\bigr)
\Bigr),\\
\mathbf{H}^{l}_{2}
=
\mathrm{LN}\Bigl(
\mathbf{H}^{l}_{1}
+
\mathrm{ACAttention}\bigl(
\mathbf{H}^{l}_{1},\mathbf{T},\mathbf{T},\hat{\mathbf{E}}^{ac}
\bigr)
\Bigr),\\
\mathbf{H}^{l}
=
\mathrm{LN}\Bigl(
\mathbf{H}^{l}_{2}
+
\mathrm{FFN}(\mathbf{H}^{l}_{2})
\Bigr),
\qquad l=1,\ldots,L_{de},
\end{gather}
where $\mathrm{LN}(\cdot)$ denotes LayerNorm \cite{ba2016layernormalization} and $\mathbf{H}^{l}\in\mathbb{R}^{\bar{s}\times d}$ is the query representation after $l$ layers.

\subsection{Meta Degradation Pattern Memory}\label{SecMemory}
Established battery knowledge \cite{attia2022knees} suggests that battery degradation trajectories share a small set of patterns across batteries.
We call these shared trajectory prototypes \emph{meta degradation patterns}, as they compose diverse real-world trajectories.
Inspired by memory networks~\cite{weston2015memorynetworks,10378627}, we propose a meta degradation pattern memory (MDPM) to store and retrieve such prototypes.
MDPM maintains $N_{\mathrm{mem}}$ learnable memory slots $\mathbf{\Omega}\in\mathbb{R}^{N_{\mathrm{mem}}\times d}$, where each slot $\mathbf{\Omega}_i\in\mathbb{R}^{d}$ stores one vector representation of a meta degradation pattern.

\textbf{Pattern retrieval.}
Given decoder output $\mathbf{H}^{L_{de}}\in\mathbb{R}^{\bar{s}\times d}$, we transform it into a memory query for retrieving relevant patterns by cosine similarity:
\begin{align}
\mathbf{q}_{mem} &= \mathrm{FFN}\left(\mathrm{Flatten}(\mathbf{H}^{L_{de}})\right)\in\mathbb{R}^{d}, \\
s_i &= \frac{\mathbf{q}_{mem}^{\top}\mathbf{\Omega}_i}{\|\mathbf{q}_{mem}\|_2\,\|\mathbf{\Omega}_i\|_2}, \qquad i=1,\ldots,N_{\mathrm{mem}}.
\end{align}
We select the top-$2$ memory slots with the largest similarity scores.
Let $\mathcal{I}_2$ denote the corresponding index set.
The relevant pattern embedding $\mathbf{h}_{mem}$ is retrieved as follows:
\begin{align}
\alpha_i &= \frac{\exp(s_i)}{\sum_{k\in\mathcal{I}_2}\exp(s_k)}, \qquad i\in\mathcal{I}_2, \\
\mathbf{h}_{mem} &= \sum_{i\in\mathcal{I}_2}\alpha_i\,\mathbf{\Omega}_i \in\mathbb{R}^{d}.
\end{align}

\textbf{Memory learning.}
During training, we encourage the retrieved pattern embedding $\mathbf{h}_{mem}$ to align with a full-life trajectory embedding $\mathbf{e}_{trajectory}$:
\begin{equation}
\mathcal{L}_{align}
=
\frac{1}{N}\sum_{i=1}^{N}
\left(
1-\frac{\mathbf{h}_{mem,i}\cdot \mathbf{e}_{trajectory,i}}{\|\mathbf{h}_{mem,i}\|_2\,\|\mathbf{e}_{trajectory,i}\|_2}
\right),
\end{equation}
where $N$ is the batch size, $\mathbf{e}_{trajectory,i}=\mathrm{TrajectoryEncoder}(\mathbf{y}_i)$, and $\mathbf{h}_{mem,i}$ is the retrieved pattern embedding for sample $i$.

To ensure $\mathbf{e}_{trajectory}$ preserves trajectory information, we reconstruct the trajectory with a decoder:
\begin{align}
\bar{\mathbf{y}} &= \mathrm{TrajectoryDecoder}(\mathbf{e}_{trajectory}), \\
\mathcal{L}_{recover}
&=
\frac{1}{N}\sum_{i=1}^{N}\frac{1}{O_i}\sum_{j=1}^{T_{\max}}
mask_{ij}\left(\mathbf{y}_{ij}-\bar{\mathbf{y}}_{ij}\right)^2,
\end{align}
where $T_{\max}=5000$ is the maximum horizon, set to cover the longest degradation trajectories in the database,
$mask_{ij}\in\{0,1\}$ indicates whether the ground-truth SOH $y_{ij}$ is available at cycle $j$ for sample $i$ and falls in the prediction region,
and $O_i=\sum_{j=1}^{T_{\max}} mask_{ij}$ is the number of observed SOH measurements.
Both $\mathrm{TrajectoryEncoder}(\cdot)$ and $\mathrm{TrajectoryDecoder}(\cdot)$ are feed-forward networks with GELU.

\textbf{Fusion and prediction.}
We incorporate the retrieved degradation pattern into the forecasting head via gated fusion:
\begin{gather}
\bar{\mathbf{H}}=
\mathrm{GELU}\left(\mathrm{Flatten}(\mathbf{H}^{L_{de}})\mathbf{W}_3+\mathbf{b}_3\right)\mathbf{W}_4+\mathbf{b}_4,\\
\boldsymbol{\beta}=
\mathrm{Sigmoid}\left(\mathrm{FFN}\left([\bar{\mathbf{H}};\mathbf{h}_{mem}]\right)\right),\\
\hat{\mathbf{y}}=
\mathrm{Head}\left(\bar{\mathbf{H}}+\boldsymbol{\beta}\odot\mathbf{h}_{mem}\right),
\end{gather}
where $\boldsymbol{\beta}\in\mathbb{R}^{d}$ is a feature-wise gate, $\odot$ denotes element-wise multiplication, and $\mathrm{Head}(\cdot)$ is a linear projection that outputs the predicted degradation trajectory $\hat{\mathbf{y}}$.

\subsection{Training of BatteryMFormer}
BatteryMFormer is trained with the following objective:
\begin{align}
\min_{\theta}\ \mathcal{L}(\theta)
&=
\mathcal{L}_{pred}+\lambda_1\mathcal{L}_{align}+\lambda_2\mathcal{L}_{recover}, \\
\mathcal{L}_{pred}
&=
\frac{1}{N}\sum_{i=1}^{N}\frac{1}{O_i}\sum_{j=1}^{T_{\max}}
mask_{ij}\bigl(\mathbf{y}_{ij}-\hat{\mathbf{y}}_{ij}\bigr)^2,
\label{loss:pred}
\end{align}
where $\lambda_1$ and $\lambda_2$ weight the alignment and recovery losses, respectively.

\begin{table}[t]
  \centering
  \caption{Statistics of four battery domains.}
  \label{tab:statistics}
  \small
  \resizebox{0.47\textwidth}{!}{\begin{tabular}{l|llll}
  \specialrule{0.1em}{0pt}{0pt}
   & Li-ion & CALB & Na-ion & Zn-ion \\
  \hline
  Number of batteries & 963 & 27 & 31 & 95 \\
  Aging conditions & 466 & 4 & 12 & 95 \\
  Charge and discharge protocols & 398 & 2 & 12 & 1 \\
  Chemical systems & 15 & 1 & 1 & 45 \\
  Operating temperatures & 8 & 4 & 1 & 3 \\
  Cycle Average & 863 & 866 & 185 & 410 \\
  Cycle Standard Deviation & 717 & 496 & 53 & 399 \\
  Cycle Minimum & 108 & 104 & 102 & 104 \\
  Cycle Maximum & 4904 & 1411  & 308 & 1652 \\
  \specialrule{0.1em}{0pt}{0pt}
  \end{tabular}}
\end{table}

\section{Experiments}

\subsection{Experimental Settings}
\textbf{Datasets}. We evaluate our model and baselines on four battery domains from the largest public real-world battery lifetime database \cite{BatteryLife}. Dataset statistics are reported in Table \ref{tab:statistics}.
\begin{itemize}[leftmargin=*, labelsep=0.5em]
    \item \textbf{Li-ion}. This domain contains lab-tested lithium-ion batteries (LIBs) aggregated from 13 subdatasets \cite{20132816492471,HE201110314,20182105222726,attia2020closed,20191406719359,juarez2020degradation,Juarez-Robles_2021,20203809192263,mohtat2021reversible,20214611166935,LI2021230024,20223412623287,zhu2022data,batteryarchive,20244517308214,wang2024physics,LI2024101891,20243216836174}. Most batteries are commercial, covering diverse operating conditions and widely used LIB chemistries.
    \item \textbf{CALB}. This domain consists of large-format commercial LIBs tested in a production environment \cite{BatteryLife}. Compared with Li-ion, CALB reflects industrial development toward larger capacities and package structure.
    \item \textbf{Na-ion}. This domain includes commercial sodium-ion batteries evaluated under diverse charge and discharge protocols \cite{BatteryLife}.
    \item \textbf{Zn-ion}. This domain contains zinc-ion batteries with varying electrolyte compositions and package structures, tested under different operating temperatures \cite{BatteryLife}.
\end{itemize}

\begin{table*}[t]
    \centering
    \caption{Overall model performance on four battery domains. The top-three results are shaded. The best results are shown in bold and the second-best results are underlined. The improvement denotes the relative improvement of BatteryMFormer over the second-best model.}
    \label{tab:overall}
    \setlength{\tabcolsep}{4pt}
    \begin{tabular}{l|cc|cc|cc|cc}
    \specialrule{0.1em}{0pt}{0pt}
    \multirow{2}{*}{Model} & \multicolumn{2}{c|}{Li-ion} & \multicolumn{2}{c|}{CALB} & \multicolumn{2}{c|}{Na-ion} & \multicolumn{2}{c}{Zn-ion} \\
    \cline{2-9}
     & MAPE (\%) & MAE ($10^{-2}$) & MAPE (\%) & MAE ($10^{-2}$) & MAPE (\%) & MAE ($10^{-2}$) & MAPE (\%) & MAE ($10^{-2}$) \\
    \hline
    PatchTST & 3.699$_{\pm0.269}$ & 3.385$_{\pm0.272}$ & 3.096$_{\pm2.703}$ & 2.906$_{\pm2.468}$ & 1.783$_{\pm0.557}$ & 1.484$_{\pm0.461}$ & 5.969$_{\pm0.630}$ & 5.773$_{\pm0.528}$ \\
    DLinear & 4.155$_{\pm0.336}$ & 3.802$_{\pm0.265}$ & 17.968$_{\pm23.386}$ & 16.429$_{\pm21.273}$ & 1.607$_{\pm0.560}$ & 1.331$_{\pm0.457}$ & 7.288$_{\pm0.821}$ & 7.029$_{\pm0.649}$ \\
    iTransformer & 2.986$_{\pm0.182}$ & 2.727$_{\pm0.190}$ & 3.203$_{\pm2.754}$ & 3.004$_{\pm2.518}$ & 1.884$_{\pm0.737}$ & 1.564$_{\pm0.605}$ & 5.938$_{\pm0.673}$ & 5.743$_{\pm0.442}$ \\
    TimesFM & 8.668$_{\pm0.311}$ & 7.680$_{\pm0.248}$ & 4.220$_{\pm1.329}$ & 3.950$_{\pm1.264}$ & 4.100$_{\pm0.731}$ & 3.378$_{\pm0.620}$ & 14.819$_{\pm0.778}$ & 13.615$_{\pm0.547}$ \\
    ConvTimeNet & 3.472$_{\pm0.236}$ & 3.157$_{\pm0.169}$ & 3.255$_{\pm2.790}$ & 3.044$_{\pm2.555}$ & 1.491$_{\pm0.653}$ & 1.233$_{\pm0.536}$ & 7.810$_{\pm0.563}$ & 7.628$_{\pm0.829}$ \\
    TimeMixer++ & 2.690$_{\pm0.135}$ & 2.446$_{\pm0.131}$ & 3.281$_{\pm2.268}$ & 3.075$_{\pm2.064}$ & 1.303$_{\pm0.461}$ & 1.082$_{\pm0.385}$ & 6.749$_{\pm0.953}$ & 6.593$_{\pm0.908}$ \\
    PatchMLP & 2.819$_{\pm0.023}$ & 2.561$_{\pm0.046}$ & 23.502$_{\pm21.780}$ & 21.888$_{\pm19.887}$ & 1.350$_{\pm0.348}$ & 1.120$_{\pm0.289}$ & 6.140$_{\pm1.254}$ & 5.835$_{\pm1.069}$ \\
    CPMLP & 2.615$_{\pm0.073}$ & 2.380$_{\pm0.067}$ & 3.371$_{\pm2.842}$ & 3.161$_{\pm2.597}$ & \cellcolor{gray!30}1.267$_{\pm0.484}$ & \cellcolor{gray!30}1.050$_{\pm0.405}$ & 6.841$_{\pm0.554}$ & 6.733$_{\pm0.623}$ \\
    CPTransformer & \cellcolor{gray!30}2.557$_{\pm0.121}$ & \cellcolor{gray!30}2.325$_{\pm0.118}$ & 3.354$_{\pm2.784}$ & 3.141$_{\pm2.549}$ & 1.511$_{\pm0.589}$ & 1.254$_{\pm0.485}$ & 5.811$_{\pm0.589}$ & 5.649$_{\pm0.512}$ \\
    TimeBridge & 3.037$_{\pm0.446}$ & 2.776$_{\pm0.438}$ & \cellcolor{gray!30}2.287$_{\pm1.296}$ & \cellcolor{gray!30}2.136$_{\pm1.186}$ & \cellcolor{gray!30}\underline{1.217$_{\pm0.517}$} & \cellcolor{gray!30}\underline{1.008$_{\pm0.436}$} & \cellcolor{gray!30}5.563$_{\pm0.403}$ & \cellcolor{gray!30}5.408$_{\pm0.219}$ \\
    IC2ML & \cellcolor{gray!30}\underline{2.528$_{\pm0.220}$} & \cellcolor{gray!30}\underline{2.284$_{\pm0.206}$} & \cellcolor{gray!30}\underline{1.955$_{\pm1.212}$} & \cellcolor{gray!30}\underline{1.892$_{\pm1.156}$} & 1.348$_{\pm0.976}$ & 1.114$_{\pm0.801}$ & \cellcolor{gray!30}\underline{5.460$_{\pm0.283}$} & \cellcolor{gray!30}\underline{5.355$_{\pm0.197}$} \\
    BatteryMFormer & \cellcolor{gray!30}\textbf{2.248$_{\pm0.037}$} & \cellcolor{gray!30}\textbf{2.034$_{\pm0.041}$} & \cellcolor{gray!30}\textbf{1.789$_{\pm0.282}$} & \cellcolor{gray!30}\textbf{1.687$_{\pm0.276}$} & \cellcolor{gray!30}\textbf{1.002$_{\pm0.446}$} & \cellcolor{gray!30}\textbf{0.830$_{\pm0.371}$} & \cellcolor{gray!30}\textbf{4.970$_{\pm0.654}$} & \cellcolor{gray!30}\textbf{4.721$_{\pm0.618}$} \\
    \hline
    Improvement & 11.07\% & 10.94\% & 8.49\% & 10.83\% & 17.66\% & 17.65\% & 8.97\% & 11.83\% \\
    \specialrule{0.1em}{0pt}{0pt}
    \end{tabular}
  \end{table*}

\begin{figure*}[t]
  \centering
  \includegraphics[width=0.86\linewidth]{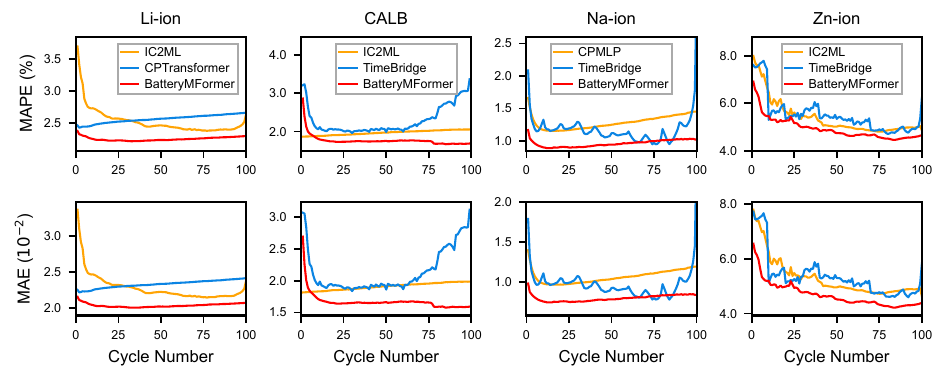}
  \caption{Performance of the top-three models as the number of usable early cycles increases.}
  \label{fig:different_cycle}
\end{figure*}
\textbf{Metrics and dataset splits}. In line with prior work~\cite{20250417745793,20242016083489}, we evaluate performance using mean absolute error (MAE) and mean absolute percentage error (MAPE), both computed on the original SOH values. We assess model generalizability under aging-condition-exclusive testing, where all test batteries come from aging conditions unseen during training and validation. For the Li-ion and Zn-ion domains, we generate three random splits while keeping the aging condition counts close to a 6:2:2 train/validation/test ratio. For CALB and Na-ion, where the number of aging conditions is limited, we use a leave-one-aging-condition-out protocol: one aging condition is held out for testing, and 25\% of the remaining aging conditions are selected for validation while the rest are used for training. We report the mean and standard deviation over the resulting splits for each domain.

\textbf{Baselines}. We compare against state-of-the-art methods in two groups.
(1) Battery-specific BDTF models: IC2ML~\cite{HUANG2026239148}, CPTransformer~\cite{BatteryLife}, and CPMLP~\cite{BatteryLife}.
(2) Generic time-series forecasting models: Transformer-based methods (TimeMixer++~\cite{20252818763501}, TimeBridge~\cite{liu2025timebridge}, iTransformer~\cite{DBLP:conf/iclr/LiuHZWWML24}, TimesFM~\cite{TimesFM}, and PatchTST~\cite{DBLP:conf/iclr/NieNSK23}), multi-layer perceptron (MLP) methods (PatchMLP~\cite{PatchMLP} and DLinear~\cite{DBLP:conf/aaai/ZengCZ023}), and convolutional neural network (CNN)-based methods (ConvTimeNet~\cite{ConvTimeNet}).
Following the BatteryLife benchmark protocol~\cite{BatteryLife}, all baselines take voltage, current, and capacity sequences as input and predict the future degradation trajectory, except IC2ML and TimesFM. Following the original designs,
IC2ML uses only the charging capacity-increment sequence as input, and TimesFM extrapolates future SOH values from the historical SOH sequence.

\textbf{Implementation details}. Following prior work~\cite{BatteryLife}, we resample each cycle to a unified length of $L=300$. All models are implemented in PyTorch and trained for up to 300 epochs with early stopping (patience 30) based on validation performance. For each trainable model and domain, we evaluate at least 10 hyperparameter configurations and report the one with the best validation performance. All experiments are conducted on NVIDIA RTX 3090 GPUs. Additional implementation and preprocessing details are provided in Appendix~\ref{Appendix:Implementation_details} and Appendix~\ref{Appendix:Data_processing}, respectively.

\subsection{Overall Performance}
  \begin{table*}[!t]
    \centering
    \caption{Ablation study of BatteryMFormer on four battery domains.}
    \label{tab:ablation}
    \setlength{\tabcolsep}{4pt}
    \renewcommand{\arraystretch}{1.1}
    \resizebox{0.925\textwidth}{!}{\begin{tabular}{l|cc|cc|cc|cc}
    \specialrule{0.1em}{0pt}{0pt}
    \multirow{2}{*}{Model} & \multicolumn{2}{c|}{Li-ion} & \multicolumn{2}{c|}{CALB} & \multicolumn{2}{c|}{Na-ion} & \multicolumn{2}{c}{Zn-ion} \\
    \cline{2-9}
     & MAPE (\%) & MAE ($10^{-2}$) & MAPE (\%) & MAE ($10^{-2}$) & MAPE (\%) & MAE ($10^{-2}$) & MAPE (\%) & MAE ($10^{-2}$) \\
    \hline
    BatteryMFormer & \textbf{2.248}$_{\pm0.037}$ & \textbf{2.034}$_{\pm0.041}$ & \textbf{1.789}$_{\pm0.282}$ & \textbf{1.687}$_{\pm0.276}$ & \textbf{1.002}$_{\pm0.446}$ & \textbf{0.830}$_{\pm0.371}$ & \textbf{4.970}$_{\pm0.654}$ & \textbf{4.721}$_{\pm0.618}$ \\
    w/o SOCView & 2.726$_{\pm0.112}$ & 2.476$_{\pm0.113}$ & 2.010$_{\pm0.681}$ & 1.898$_{\pm0.615}$ & 1.393$_{\pm0.399}$ & 1.156$_{\pm0.331}$ & 6.647$_{\pm1.309}$ & 6.343$_{\pm1.193}$ \\
    w/o MDPM & 2.497$_{\pm0.172}$ & 2.268$_{\pm0.168}$ & 2.158$_{\pm0.878}$ & 2.031$_{\pm0.788}$ & 1.252$_{\pm0.459}$ & 1.038$_{\pm0.383}$ & 6.133$_{\pm1.705}$ & 5.924$_{\pm1.597}$ \\
    w/o ACDecoder & 2.560$_{\pm0.042}$ & 2.328$_{\pm0.048}$ & 2.449$_{\pm1.334}$ & 2.301$_{\pm1.209}$ & 1.243$_{\pm0.422}$ & 1.029$_{\pm0.351}$ & 7.722$_{\pm0.384}$ & 7.516$_{\pm0.601}$ \\
    w/o ACAttention & 2.488$_{\pm0.045}$ & 2.261$_{\pm0.035}$ & 2.058$_{\pm0.743}$ & 1.941$_{\pm0.670}$ & 1.229$_{\pm0.401}$ & 1.018$_{\pm0.332}$ & 7.602$_{\pm1.009}$ & 7.446$_{\pm1.181}$ \\
    w/o ACQuery & 2.540$_{\pm0.060}$ & 2.306$_{\pm0.040}$ & 2.263$_{\pm1.007}$ & 2.129$_{\pm0.905}$ & 1.226$_{\pm0.376}$ & 1.016$_{\pm0.314}$ & 6.324$_{\pm1.369}$ & 6.095$_{\pm1.299}$ \\
    w/o LLM & 2.270$_{\pm0.167}$ & 2.056$_{\pm0.157}$ & 3.151$_{\pm2.938}$ & 2.948$_{\pm2.693}$ & 1.011$_{\pm0.472}$ & 0.839$_{\pm0.395}$ & 5.614$_{\pm0.940}$ & 5.369$_{\pm0.896}$ \\
    \hline
    CPTransformer-SI & 2.629$_{\pm0.173}$ & 2.391$_{\pm0.170}$ & 3.162$_{\pm2.641}$ & 2.957$_{\pm2.419}$ & 1.569$_{\pm0.517}$ & 1.302$_{\pm0.430}$ & 5.581$_{\pm0.436}$ & 5.358$_{\pm0.366}$ \\
    \specialrule{0.1em}{0pt}{0pt}
    \end{tabular}}
  \end{table*}

\subsubsection{Main results}
Table \ref{tab:overall} compares BatteryMFormer with state-of-the-art baselines across four battery domains. BatteryMFormer achieves the best performance on all domains and metrics despite substantial differences in battery chemistry, data scale, and degradation characteristics. Compared with the second-best model in each domain, BatteryMFormer reduces MAPE by 11.07\%, 8.49\%, 17.66\%, and 8.97\%, and reduces MAE by 10.94\%, 10.83\%, 17.65\%, and 11.83\% on Li-ion, CALB, Na-ion, and Zn-ion, respectively. These consistent improvements demonstrate the effectiveness of BatteryMFormer for early BDTF across domains.

Notably, the strongest baseline varies across domains: IC2ML performs best among baselines on Li-ion, CALB, and Zn-ion, while TimeBridge achieves the best baseline performance on Na-ion. This suggests that the predictive patterns underlying battery degradation are domain-dependent, and different architectural inductive biases match these patterns to different extents. This heterogeneity regarding the underlying effective patterns can also lead to pronounced performance instability. For example, DLinear performs competitively on Na-ion but incurs much larger errors on CALB, and TimesFM shows relatively high errors on Li-ion, Na-ion and Zn-ion. In contrast, BatteryMFormer consistently achieves the best performance across domains, indicating that it can capture a broader spectrum of degradation patterns in different battery datasets.

\subsubsection{Comparison under different numbers of usable cycles}
We further evaluate model performance by varying the number of usable early cycles. Figure \ref{fig:different_cycle} reports MAE and MAPE for the top-performing models on each domain. BatteryMFormer achieves consistent improvements across a broad range of early-forecasting settings across four domains. These results demonstrate the effectiveness of BatteryMFormer under different amounts of early degradation information.

We also observe that prediction errors can increase on Li-ion and Na-ion when $S>25$. This reflects an open challenge in long-sequence time-series modeling: longer inputs do not guarantee improvements, and may instead introduce redundancy and optimization difficulty. In our setting, each cycle contains 300 points, so $S>25$ already yields more than 7{,}500 input points. Since adjacent battery cycles often change only marginally, longer inputs may dilute informative degradation signatures with redundant measurements. Similar error increases are also observed in other top-performing baselines and broader time-series forecasting studies when modeling long input sequences~\cite{DBLP:conf/iclr/NieNSK23,PatchMLP}. Despite this effect, BatteryMFormer generally outperforms the baselines, underscoring the advantage of the proposed multi-level learning strategy in handling long operational voltage and current time series for early BDTF.

\subsection{Ablation Study}
We conduct an ablation study of BatteryMFormer to evaluate the effectiveness of its key components, with the results summarized in Table \ref{tab:ablation}. "w/o SOCView" removes the SOC view from the dual-view encoder, and "w/o MDPM" removes the meta degradation pattern memory. For the aging-condition-aware decoder, "w/o ACQuery" removes aging-condition information from generic queries, "w/o ACAttention" replaces aging-condition-aware attention with standard attention, and "w/o ACDecoder" removes both mechanisms. "w/o LLM" replaces the language-based embedder with factor-wise lookup embeddings followed by projection. Since variable-length protocols, such as multi-stage charge/discharge protocols, cannot be trivially encoded by fixed lookup embeddings due to out-of-vocabulary issues, this variant only uses positive electrode, negative electrode, operating temperature, package structure, and manufacturer as lookup-embedding factors. "CPTransformer-SI" provides CPTransformer with the same input information as BatteryMFormer, including voltage, current, capacity, SOC, aging-condition information, and cycle-level descriptors.

Table \ref{tab:ablation} shows that the three major components of BatteryMFormer all contribute to performance. Removing the SOC view, MDPM, or ACDecoder consistently degrades results across all domains, indicating that learning SOC-localized patterns, trajectory-level prototypes, and aging-condition-informed representations is useful for early BDTF. Their contributions are domain-dependent: the SOC view brings particularly clear gains on Li-ion and Na-ion, while ACDecoder has more pronounced effects on CALB and Zn-ion. This suggests that the effective predictive patterns vary across battery domains, and that the components of BatteryMFormer capture complementary aspects of these patterns. CPTransformer-SI remains clearly worse than BatteryMFormer and performs close to CPTransformer in most domains, indicating that simply providing the same input information yields limited benefits. The advantage of BatteryMFormer therefore comes primarily from its multi-level learning architecture rather than additional input variables alone. Finer-grained ACDecoder ablations further show that both ACQuery and ACAttention are important for aging-condition-aware pattern mining. Finally, replacing the LLM embedder with lookup embeddings consistently degrades performance, confirming that semantic aging-condition representations provide useful information beyond factor-wise embeddings. While the mean difference is small on Li-ion and Na-ion, the larger standard deviations of w/o LLM suggest that the LLM embedder improves performance stability; on CALB and Zn-ion, it improves both accuracy and robustness. Collectively, these results validate the effectiveness of the proposed multi-level learning strategy.

\begin{figure}[t]
\centering
\includegraphics  
[width=\linewidth]{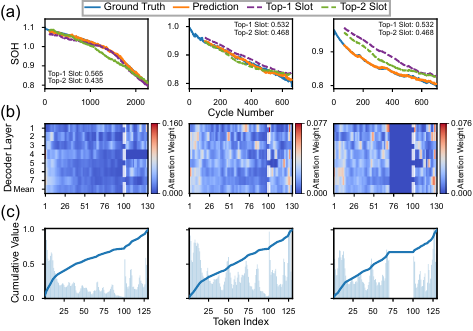}
\caption{Case study on three representative test batteries with superlinear, linear, and sublinear degradation. (a) Ground-truth vs.\ predicted SOH trajectories with the top two retrieved MDPM prototypes (weights shown). (b) Attention weights of ACDecoder cross-attention over temporal-view and SOC-view tokens; dashed line marks the boundary. (c) Token-wise attention weights (bars) and cumulative weight (lines) over token indices.}
\label{fig:case_study}
\end{figure}

\begin{figure}[t]
    \centering
    \includegraphics[width=\linewidth]{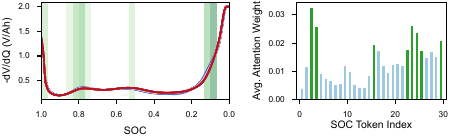}
\caption{Differential voltage analysis and average attention weights of SOC-view tokens for a representative test battery. Highlighted regions denote the SOC intervals corresponding to the top-25\% SOC-view tokens ranked by attention weight, with darker green indicating larger attention weights.}
    \label{fig:dva_attention}
\end{figure}
\subsection{Case Study}
To interpret how BatteryMFormer leverages the multi-level structure of early BDTF, we examine three representative test batteries exhibiting superlinear, linear, and sublinear degradation (Figure~\ref{fig:case_study}). We analyze (i) the top-two meta degradation patterns retrieved from MDPM, visualized by decoding the corresponding memory embeddings with the trajectory decoder, and (ii) the ACDecoder cross-attention weights over temporal-view and SOC-view tokens.

Figure \ref{fig:case_study}a shows that MDPM retrieves prototypes that are consistent with the batteries' long-horizon degradation patterns. For the superlinear battery, the retrieved prototypes include two trajectory prototypes showing accelerated degradation with knee points, providing informative priors for long-range extrapolation. For the linear battery, the retrieved prototypes are approximately linear and largely agree with the observed trend. For the sublinear battery, MDPM retrieves prototypes exhibiting a slowdown in degradation even though the input covers only the early, faster-decay stage, indicating that the MDPM stores diverse global trajectory shapes and can retrieve related trajectory prototypes to help BDTF for batteries subjected to aging conditions not covered by the training data.

Figure \ref{fig:case_study}b--c further indicates that ACDecoder integrates both views. Most attention mass is assigned to temporal-view tokens, while SOC-view tokens receive a non-trivial share. Moreover, attention over SOC-view tokens is highly concentrated on a small subset, suggesting that degradation-relevant operational signatures are localized to specific SOC intervals and that BatteryMFormer can prioritize these regions through attention. Overall, this case study illustrates how MDPM supplies pattern-level priors and how the dual-view encoder with ACDecoder selectively aggregates temporal and SOC-localized cues to improve early BDTF.

We further interpret the SOC-view attention through differential voltage analysis (DVA) on a test battery (Figure~\ref{fig:dva_attention}). The top-25\% SOC tokens ranked by attention weights are concentrated around the major DVA peaks and their shoulder regions, which are known to be sensitive to degradation-induced changes such as peak shifts, broadening, and shape distortion caused by lithium inventory loss, loss of active material, or polarization growth~\cite{BIRKL2017373,TAN2024103725}. This result suggests that the SOC view guides BatteryMFormer to selectively attend to particular SOC intervals. The alignment with DVA features further indicates that the learned SOC-localized patterns can reflect electrochemical signatures associated with battery aging mechanisms.
\begin{table}[t]
\centering
\caption{Results of data-efficient learning with 50\% of the training batteries retained. Top-3 and Top-2 denote the third-best and second-best baselines selected from the overall comparison.}
\label{tab:few_shot}
\small
\setlength{\tabcolsep}{5pt}
\resizebox{0.47\textwidth}{!}{\begin{tabular}{l|l|lll}
\specialrule{0.1em}{0pt}{0pt}
\multicolumn{1}{l|}{Dataset} & Metrics & Top-3 & Top-2 & BatteryMFormer \\ \hline
\multirow{2}{*}{Li-ion} & MAPE (\%)       & 2.958$_{\pm0.134}$ & 2.802$_{\pm0.320}$ & \textbf{2.453$_{\pm0.096}$} \\
                            & MAE ($10^{-2}$) & 2.696$_{\pm0.104}$ & 2.530$_{\pm0.297}$ & \textbf{2.222$_{\pm0.077}$} \\ \hline
\multirow{2}{*}{CALB}   & MAPE (\%)       & 2.426$_{\pm1.072}$ & 2.237$_{\pm1.573}$ & \textbf{2.174$_{\pm0.905}$} \\
                            & MAE ($10^{-2}$) & 2.272$_{\pm0.986}$ & 2.162$_{\pm1.511}$ & \textbf{2.049$_{\pm0.832}$} \\ \hline
\multirow{2}{*}{Na-ion} & MAPE (\%)       & 1.497$_{\pm0.538}$ & 1.234$_{\pm0.360}$ & \textbf{1.046$_{\pm0.455}$} \\
                            & MAE ($10^{-2}$) & 1.245$_{\pm0.448}$ & 1.021$_{\pm0.296}$ & \textbf{0.868$_{\pm0.381}$} \\ \hline
\multirow{2}{*}{Zn-ion} & MAPE (\%)       & 7.742$_{\pm1.115}$ & 7.545$_{\pm0.699}$ & \textbf{6.210$_{\pm0.334}$} \\
                            & MAE ($10^{-2}$) & 7.426$_{\pm0.873}$ & 7.402$_{\pm1.040}$ & \textbf{6.066$_{\pm0.204}$} \\
\specialrule{0.1em}{0pt}{0pt}
\end{tabular}}
\end{table}

\subsection{Data-Efficient Learning}
Collecting full-life degradation trajectories is costly and can take months to years, making data-efficient BDTF an important practical requirement. To evaluate model robustness under limited lifetime data, we retain only 50\% of the training batteries in each domain while keeping the validation and test parts unchanged. Table \ref{tab:few_shot} reports the results of BatteryMFormer and the top-performing baselines under this reduced-data setting.

BatteryMFormer achieves the best performance across all four domains with only 50\% of the training data. Compared with the strongest baseline, it reduces MAPE by 12.45\%, 2.81\%, 15.23\%, and 17.69\%, and reduces MAE by 12.17\%, 5.22\%, 14.98\%, and 18.04\% on Li-ion, CALB, Na-ion, and Zn-ion, respectively. The improvements are particularly pronounced on Na-ion and Zn-ion, where training data are limited and aging conditions are diverse. These results indicate that the proposed multi-level learning strategy can still extract informative degradation patterns from limited lifetime data, thereby improving data efficiency in early BDTF.

\section{Related Work}
We review existing approaches on battery degradation trajectory forecasting (BDTF) from two perspectives: feature engineering and representation learning.

\textbf{Feature-engineering-based methods.}
These methods focus on extracting degradation-relevant descriptors from operational measurements (e.g., voltage, current, capacity, relaxation signals) and then fit data-driven predictors for future capacity/SOH trajectories \cite{20250417745793,10332202,10491306}. A common practice is to extract features from regions where aging signatures are pronounced. For instance, \cite{10332202} constructs descriptors from the late-charge capacity sequence and the post-charge relaxation voltage region and then uses an LSTM as the forecaster; \cite{20250417745793} designs features tailored to a 9-step charging protocol and applies a feed-forward network for trajectory prediction. While effective on curated settings, these handcrafted descriptors are often protocol- or dataset-specific (e.g., tied to particular voltage windows or multi-step procedures) and may not be available or predictive across diverse aging conditions \cite{20223412623287,BatteryLife}.

\textbf{Representation-learning-based methods.}
In contrast, methods in this research line learn forecasting-relevant representations directly from raw or minimally processed measurements using neural networks \cite{LI2021230024,LI2022453,TAN2024103725,LIU2025114736,HUANG2026239148,BatteryLife,HUANG2024122825}. An intuitive strategy treats BDTF as generic time-series forecasting by extrapolating future SOH from historical SOH records using architectures such as LSTM or Transformer variants \cite{LI2021230024,TAN2024103725,20254519447317,LI2022453}. However, SOH-only inputs can be weakly informative in early cycles, where early trajectories appear similar yet diverge substantially in the long horizon. Recent work therefore exploits fine-grained voltage-current profiles. The BatteryLife benchmark \cite{BatteryLife} shows that directly applying generic forecasters for modeling voltage and current time series can be suboptimal, and introduces CyclePatch to model intra-cycle dynamics and inter-cycle evolution more effectively; IC2ML \cite{HUANG2026239148} further improves this paradigm by injecting auxiliary supervision to enhance learned representations. Our BatteryMFormer belongs to this representation-learning family, but advances beyond existing approaches by explicitly integrating multi-level inductive biases for early BDTF.

\section{Limitations and Ethical Considerations}
\textbf{Limitations.}
First, using more early cycles yields long and redundant inputs; for example, more than 25 cycles already corresponds to over 7,500 input points. This can compromise model performance on ultra-long inputs.
Second, we evaluate on regular laboratory/production tests that are critical for battery optimization and production, whereas field data (e.g., EV logs) are often irregular and noisier due to varying usage and sensor noise. Applying BatteryMFormer to field conditions may require modified representations and preprocessing (e.g., handling inaccurate and irregular data records \cite{Warpformer}).

\textbf{Ethical considerations.}
This work utilizes publicly available battery lifetime datasets and contains no human-subject or personally identifiable data. In field deployment, forecast errors can induce suboptimal decisions (e.g., premature retirement or delayed maintenance), so models should be validated against the target operating distribution before high-stakes deployment.

\section{Conclusion and Future Work}
This paper highlights the value of explicitly modeling the multi-level structure in early battery degradation trajectory forecasting, spanning trajectory patterns, aging conditions, and battery-specific dynamics. We propose BatteryMFormer and demonstrate that our model delivers consistent improvements over state-of-the-art baselines across four battery domains. Ablation and case studies further confirm that each component contributes meaningfully to these gains. BatteryMFormer also performs better in data-efficient settings under reduced training data. Future endeavors will focus on improving the model’s ability to model long operational time series and adapting the framework to irregular, noisy field data.

\section{GenAI Disclosure}
Generative AI tools were used to assist with language editing (e.g., improving clarity, grammar, and conciseness) of author-written text and to support code development (e.g., drafting or refactoring implementation snippets). These tools were not used to generate the experimental results reported in this work. All AI-assisted edits to the manuscript and code were reviewed and validated by the authors, who take full responsibility for the correctness, originality, and integrity of the work.

\section{Acknowledgments}
The authors acknowledge financial support from the National Key R\&D Program of China (No. 2023YFB2503600). This work is also supported by research grants from the National Natural Science Foundation of China (Nos. 92372109, 62572418 and 52207230) and the Guangdong Provincial Talent Program (No. 2024TQ08X366). We also acknowledge support from the Wilson Tang Brilliant Energy Science and Technology Lab (BEST Lab) at The Hong Kong University of Science and Technology (Guangzhou).
\bibliographystyle{ACM-Reference-Format}
\bibliography{sample-base}

@String{Computing = "Computing" }

@String{Computer = "{IEEE} Computer" }

@article{LI2024101891,
title = {Predicting battery lifetime under varying usage conditions from early aging data},
journal = {Cell Reports Physical Science},
volume = {5},
number = {4},
pages = {101891},
year = {2024},
issn = {2666-3864},
doi = {https://doi.org/10.1016/j.xcrp.2024.101891},
url = {https://www.sciencedirect.com/science/article/pii/S2666386424001279},
author = {Tingkai Li and Zihao Zhou and Adam Thelen and David A. Howey and Chao Hu}
}

@article{zhu2022data,
  title={Data-driven capacity estimation of commercial lithium-ion batteries from voltage relaxation},
  author={Zhu, Jiangong and Wang, Yixiu and Huang, Yuan and Bhushan Gopaluni, R and Cao, Yankai and Heere, Michael and M{\"u}hlbauer, Martin J and Mereacre, Liuda and Dai, Haifeng and Liu, Xinhua and others},
  journal={Nature communications},
  volume={13},
  number={1},
  pages={2261},
  year={2022},
  publisher={Nature Publishing Group UK London}
}

@article{LIU2025114736,
title = {Physics-guided TL-LSTM network for early-stage degradation trajectory prediction of lithium-ion batteries},
journal = {Journal of Energy Storage},
volume = {106},
pages = {114736},
year = {2025},
issn = {2352-152X},
doi = {https://doi.org/10.1016/j.est.2024.114736},
url = {https://www.sciencedirect.com/science/article/pii/S2352152X24043226},
author = {Qingqiang Liu and Zhiqing Shang and Shixiang Lu and Yuanhong Liu and Yuchao Liu and Sheng Yu}
}

@article{HUANG2024122825,
title = {A transferable long-term lithium-ion battery aging trajectory prediction model considering internal resistance and capacity regeneration phenomenon},
journal = {Applied Energy},
volume = {360},
pages = {122825},
year = {2024},
issn = {0306-2619},
doi = {https://doi.org/10.1016/j.apenergy.2024.122825},
url = {https://www.sciencedirect.com/science/article/pii/S0306261924002083},
author = {Yaodi Huang and Pengcheng Zhang and Jiahuan Lu and Rui Xiong and Zhongmin Cai}
}

@article{attia2022knees,
  title={“Knees” in lithium-ion battery aging trajectories},
  author={Attia, Peter M and Bills, Alexander and Planella, Ferran Brosa and Dechent, Philipp and Dos Reis, Goncalo and Dubarry, Matthieu and Gasper, Paul and Gilchrist, Richard and Greenbank, Samuel and Howey, David and others},
  journal={Journal of The Electrochemical Society},
  volume={169},
  number={6},
  pages={060517},
  year={2022},
  publisher={IOP Publishing}
}

@ARTICLE{10491306,
  author={Meng, Jinhao and Cai, Lei and Yang, Shengxiang and Li, Junxin and Zhou, Feifan and Peng, Jichang and Song, Zhengxiang},
  journal={IEEE Transactions on Energy Conversion}, 
  title={An Empirical-Informed Model for the Early Degradation Trajectory Prediction of Lithium-ion Battery}, 
  year={2024},
  volume={39},
  number={4},
  pages={2299-2311},
  doi={10.1109/TEC.2024.3385093}}

@article{HE201110314,
title = {Prognostics of lithium-ion batteries based on Dempster–Shafer theory and the Bayesian Monte Carlo method},
journal = {Journal of Power Sources},
volume = {196},
number = {23},
pages = {10314-10321},
year = {2011},
issn = {0378-7753},
doi = {https://doi.org/10.1016/j.jpowsour.2011.08.040},
url = {https://www.sciencedirect.com/science/article/pii/S0378775311015400},
author = {Wei He and Nicholas Williard and Michael Osterman and Michael Pecht}
}

@article{tan2025PBT,
  title={Pretrained Battery Transformer (PBT): A battery life prediction foundation model},
  author={Tan, Ruifeng and Hong, Weixiang and Li, Jia and Huang, Jiaqiang and Zhang, Tong-Yi},
  journal={arXiv preprint arXiv:2512.16334},
  year={2025}
}

@article{zhang2025unlocking,
  title={Unlocking Ultrafast Diagnosis of Retired Batteries via Interpretable Machine Learning and Optical Fiber Sensors},
  author={Zhang, Taolue and Tan, Ruifeng and Zhu, Pinxi and Zhang, Tong-Yi and Huang, Jiaqiang},
  journal={ACS Energy Letters},
  volume={10},
  pages={862--871},
  year={2025},
  publisher={ACS Publications}
}

@article{fleischmann2023battery,
  title={Battery 2030: Resilient, sustainable, and circular},
  author={Fleischmann, Jakob and Hanicke, Mikael and Horetsky, Evan and Ibrahim, Dina and Jautelat, S{\"o}ren and Linder, Martin and Schaufuss, Patrick and Torscht, Lukas and van de Rijt, Alexandre},
  journal={McKinsey \& Company},
  volume={16},
  pages={2023},
  year={2023}
}

@inproceedings{LiPM,
author = {Li, Juren and Yang, Yang and Su, Hanchen and Liu, Jiayu and Chen, Youmin and Zhang, Jianfeng and Pan, Lujia},
title = {LiPM: Foundation Model for Lithium-Ion Battery Analysis},
year = {2025},
isbn = {9798400714542},
publisher = {Association for Computing Machinery},
address = {New York, NY, USA},
url = {https://doi.org/10.1145/3711896.3737027},
doi = {10.1145/3711896.3737027},
booktitle = {Proceedings of the 31st ACM SIGKDD Conference on Knowledge Discovery and Data Mining V.2},
pages = {1412–1423},
numpages = {12},
location = {Toronto ON, Canada},
series = {KDD '25}
}

@article{HUANG2026239148,
title = {IC2ML: Unified battery state-of-health, degradation trajectory and remaining useful life prediction via intra-cycle and inter-cycle enhanced machine learning},
journal = {Journal of Power Sources},
volume = {666},
pages = {239148},
year = {2026},
issn = {0378-7753},
doi = {https://doi.org/10.1016/j.jpowsour.2025.239148},
url = {https://www.sciencedirect.com/science/article/pii/S0378775325029854},
author = {Xinghao Huang and Chen Liang and Shengyu Tao and Yunhong Che and Ningyu Bian and Jiale Zhang and Runhua Wang and Yuqi Zhang and Bizhong Xia and Xuan Zhang}
}

@misc{ba2016layernormalization,
      title={Layer Normalization}, 
      author={Jimmy Lei Ba and Jamie Ryan Kiros and Geoffrey E. Hinton},
      year={2016},
      eprint={1607.06450},
      archivePrefix={arXiv},
      primaryClass={stat.ML},
      url={https://arxiv.org/abs/1607.06450}, 
}

@article{LI2021230024,
title = {One-shot battery degradation trajectory prediction with deep learning},
journal = {Journal of Power Sources},
volume = {506},
pages = {230024},
year = {2021},
issn = {0378-7753},
doi = {https://doi.org/10.1016/j.jpowsour.2021.230024},
url = {https://www.sciencedirect.com/science/article/pii/S0378775321005528},
author = {Weihan Li and Neil Sengupta and Philipp Dechent and David Howey and Anuradha Annaswamy and Dirk Uwe Sauer}
}

@inproceedings{20182105216120,
language = {English},
copyright = {Compilation and indexing terms, Copyright 2025 Elsevier Inc.},
copyright = {Compendex},
title = {Attention is all you need},
journal = {Advances in Neural Information Processing Systems},
author = {Vaswani, Ashish and Shazeer, Noam and Parmar, Niki and Uszkoreit, Jakob and Jones, Llion and Gomez, Aidan N. and Kaiser, Lukasz and Polosukhin, Illia},
volume = {2017-December},
booktitle = {Advances in Neural Information Processing Systems},
year = {2017},
pages = {5999 - 6009},
issn = {10495258},
address = {Long Beach, CA, United states}
}

@article{BIRKL2017373,
title = {Degradation diagnostics for lithium ion cells},
journal = {Journal of Power Sources},
volume = {341},
pages = {373-386},
year = {2017},
issn = {0378-7753},
doi = {https://doi.org/10.1016/j.jpowsour.2016.12.011},
url = {https://www.sciencedirect.com/science/article/pii/S0378775316316998},
author = {Christoph R. Birkl and Matthew R. Roberts and Euan McTurk and Peter G. Bruce and David A. Howey}
}

@inproceedings{DBLP:conf/iclr/LiuHZWWML24,
  author       = {Yong Liu and
                  Tengge Hu and
                  Haoran Zhang and
                  Haixu Wu and
                  Shiyu Wang and
                  Lintao Ma and
                  Mingsheng Long},
  title        = {iTransformer: Inverted Transformers Are Effective for Time Series
                  Forecasting},
  booktitle    = {The Twelfth International Conference on Learning Representations,
                  {ICLR} 2024, Vienna, Austria, May 7-11, 2024},
  publisher    = {OpenReview.net},
  year         = {2024},
  pages = {1-25},
  url          = {https://openreview.net/forum?id=JePfAI8fah},
  bibsource    = {dblp computer science bibliography, https://dblp.org},
  address = {Vienna, Austria},
}

@inproceedings{20252818763501,
language = {English},
copyright = {Compilation and indexing terms, Copyright 2026 Elsevier Inc.},
copyright = {Compendex},
title = {TIMEMIXER++: A GENERAL TIME SERIES PATTERN MACHINE FOR UNIVERSAL PREDICTIVE ANALYSIS},
journal = {13th International Conference on Learning Representations, ICLR 2025},
author = {Wang, Shiyu and Li, Jiawei and Shi, Xiaoming and Ye, Zhou and Mo, Baichuan and Lin, Wenze and Ju, Shengtong and Chu, Zhixuan and Jin, Ming},
year = {2025},
pages = {1662 - 1698},
address = {Singapore, Singapore}
}

@inproceedings{DBLP:conf/aaai/ZengCZ023,
  author       = {Ailing Zeng and
                  Muxi Chen and
                  Lei Zhang and
                  Qiang Xu},
  editor       = {Brian Williams and
                  Yiling Chen and
                  Jennifer Neville},
  title        = {Are Transformers Effective for Time Series Forecasting?},
  booktitle    = {Thirty-Seventh {AAAI} Conference on Artificial Intelligence, {AAAI}
                  2023, Thirty-Fifth Conference on Innovative Applications of Artificial
                  Intelligence, {IAAI} 2023, Thirteenth Symposium on Educational Advances
                  in Artificial Intelligence, {EAAI} 2023, Washington, DC, USA, February
                  7-14, 2023},
  address = {Washington, DC, USA},
  pages        = {11121--11128},
  publisher    = {{AAAI} Press},
  year         = {2023},
  url          = {https://doi.org/10.1609/aaai.v37i9.26317},
  doi          = {10.1609/AAAI.V37I9.26317},
  bibsource    = {dblp computer science bibliography, https://dblp.org}
}

@article{Juarez-Robles_2021,
doi = {10.1149/1945-7111/ac30af},
url = {https://dx.doi.org/10.1149/1945-7111/ac30af},
year = {2021},
month = {nov},
publisher = {IOP Publishing},
volume = {168},
number = {11},
pages = {110501},
author = {Juarez-Robles, Daniel and Azam, Saad and Jeevarajan, Judith A. and Mukherjee, Partha P.},
title = {Degradation-Safety Analytics in Lithium-Ion Cells and Modules: Part III. Aging and Safety of Pouch Format Cells},
journal = {Journal of The Electrochemical Society}
}

@article{tao2023collaborative,
  title={Collaborative and privacy-preserving retired battery sorting for profitable direct recycling via federated machine learning},
  author={Tao, Shengyu and Liu, Haizhou and Sun, Chongbo and Ji, Haocheng and Ji, Guanjun and Han, Zhiyuan and Gao, Runhua and Ma, Jun and Ma, Ruifei and Chen, Yuou and others},
  journal={Nature Communications},
  volume={14},
  number={1},
  pages={8032},
  year={2023},
  publisher={Nature Publishing Group UK London}
}

@inproceedings{DBLP:conf/iclr/NieNSK23,
  author       = {Yuqi Nie and
                  Nam H. Nguyen and
                  Phanwadee Sinthong and
                  Jayant Kalagnanam},
  title        = {A Time Series is Worth 64 Words: Long-term Forecasting with Transformers},
  pages = {1-24},
  booktitle    = {The Eleventh International Conference on Learning Representations, {ICLR} 2023, Kigali, Rwanda, May 1-5, 2023},
  publisher    = {OpenReview.net},
  address = {Kigali, Rwanda},
  year         = {2023},
  url          = {https://openreview.net/forum?id=Jbdc0vTOcol},
  bibsource    = {dblp computer science bibliography, https://dblp.org}
}

@article{PatchMLP, title={Unlocking the Power of Patch: Patch-Based MLP for Long-Term Time Series Forecasting}, volume={39}, url={https://ojs.aaai.org/index.php/AAAI/article/view/33378}, DOI={10.1609/aaai.v39i12.33378}, number={12}, journal={Proceedings of the AAAI Conference on Artificial Intelligence}, author={Tang, Peiwang and Zhang, Weitai}, year={2025}, month={Apr.}, pages={12640-12648} }

@article{qwen3embedding,
  title={Qwen3 Embedding: Advancing Text Embedding and Reranking Through Foundation Models},
  author={Zhang, Yanzhao and Li, Mingxin and Long, Dingkun and Zhang, Xin and Lin, Huan and Yang, Baosong and Xie, Pengjun and Yang, An and Liu, Dayiheng and Lin, Junyang and Huang, Fei and Zhou, Jingren},
  journal={arXiv preprint arXiv:2506.05176},
  year={2025}
}

@inproceedings{TimesFM,
author = {Das, Abhimanyu and Kong, Weihao and Sen, Rajat and Zhou, Yichen},
title = {A decoder-only foundation model for time-series forecasting},
year = {2024},
publisher = {JMLR.org},
booktitle = {Proceedings of the 41st International Conference on Machine Learning},
articleno = {404},
numpages = {20},
location = {Vienna, Austria},
series = {ICML'24}
}

@inproceedings{ConvTimeNet,
author = {Cheng, Mingyue and Yang, Jiqian and Pan, Tingyue and Liu, Qi and Li, Zhi and Wang, Shijin},
title = {ConvTimeNet: A Deep Hierarchical Fully Convolutional Model for Multivariate Time Series Analysis},
year = {2025},
isbn = {9798400713316},
publisher = {Association for Computing Machinery},
address = {New York, NY, USA},
url = {https://doi.org/10.1145/3701716.3715214},
doi = {10.1145/3701716.3715214},
booktitle = {Companion Proceedings of the ACM on Web Conference 2025},
pages = {171–180},
numpages = {10},
location = {Sydney NSW, Australia},
series = {WWW '25}
}

@article{zhang2025battery,
Author = {Zhang, Han and Li, Yuqi and Zheng, Shun and Lu, Ziheng and Gui, Xiaofan
   and Xu, Wei and Bian, Jiang},
Title = {Battery lifetime prediction across diverse ageing conditions with
   inter-cell deep learning},
Journal = {Nature Machine Intelligence},
Year = {2025},
Volume = {7},
Number = {2},
Pages = {270-277},
Month = {FEB},
DOI = {10.1038/s42256-024-00972-x},
EarlyAccessDate = {JAN 2025},
EISSN = {2522-5839},
ResearcherID-Numbers = {LU, Ziheng/N-4349-2018
   Li, Yuqi/R-3015-2019
   Gui, Xiaofan/KIB-1146-2024
   },
ORCID-Numbers = {Bian, Jiang/0000-0002-9472-600X
   Li, Yuqi/0000-0003-1501-1549
   zheng, shun/0009-0005-7355-7090
   Gui, Xiaofan/0000-0001-8259-5471},
Unique-ID = {WOS:001396148200001},
}

@article{juarez2020degradation,
  title={Degradation-safety analytics in lithium-ion cells: Part I. Aging under charge/discharge cycling},
  author={Juarez-Robles, Daniel and Jeevarajan, Judith A and Mukherjee, Partha P},
  journal={Journal of The Electrochemical Society},
  volume={167},
  number={16},
  pages={160510},
  year={2020},
  publisher={IOP Publishing}
}

@article{20132816492471 ,
language = {English},
copyright = {Compilation and indexing terms, Copyright 2025 Elsevier Inc.},
copyright = {Compendex},
title = {An ensemble model for predicting the remaining useful performance of lithium-ion batteries},
journal = {Microelectronics Reliability},
author = {Xing, Yinjiao and Ma, Eden W.M. and Tsui, Kwok-Leung and Pecht, Michael},
volume = {53},
number = {6},
year = {2013},
pages = {811 - 820},
issn = {00262714},
URL = {http://dx.doi.org/10.1016/j.microrel.2012.12.003},
}

@article{TAN2024103725,
title = {Forecasting battery degradation trajectory under domain shift with domain generalization},
journal = {Energy Storage Materials},
volume = {72},
pages = {103725},
year = {2024},
issn = {2405-8297},
doi = {https://doi.org/10.1016/j.ensm.2024.103725},
url = {https://www.sciencedirect.com/science/article/pii/S2405829724005518},
author = {Ruifeng Tan and Xibin Lu and Minhao Cheng and Jia Li and Jiaqiang Huang and Tong-Yi Zhang}
}

@article{attia2020closed,
  title={Closed-loop optimization of fast-charging protocols for batteries with machine learning},
  author={Attia, Peter M and Grover, Aditya and Jin, Norman and Severson, Kristen A and Markov, Todor M and Liao, Yang-Hung and Chen, Michael H and Cheong, Bryan and Perkins, Nicholas and Yang, Zi and others},
  journal={Nature},
  volume={578},
  number={7795},
  pages={397--402},
  year={2020},
  publisher={Nature Publishing Group UK London}
}

@article{20223412623287 ,
language = {English},
copyright = {Compilation and indexing terms, Copyright 2024 Elsevier Inc.},
copyright = {Compendex},
title = {Real-time personalized health status prediction of lithium-ion batteries using deep transfer learning},
journal = {Energy and Environmental Science},
author = {Ma, Guijun and Xu, Songpei and Jiang, Benben and Cheng, Cheng and Yang, Xin and Shen, Yue and Yang, Tao and Huang, Yunhui and Ding, Han and Yuan, Ye},
volume = {15},
number = {10},
year = {2022},
pages = {4083 - 4094},
issn = {17545692},
URL = {http://dx.doi.org/10.1039/d2ee01676a}
}

@unpublished{20243216836174,
language = {English},
copyright = {Compilation and indexing terms, Copyright 2025 Elsevier Inc.},
copyright = {Compendex},
title = {BATTERYML: AN OPEN-SOURCE PLATFORM FOR MACHINE LEARNING ON BATTERY DEGRADATION},
journal = {arXiv},
author = {Zhang, Han and Gui, Xiaofan and Zheng, Shun and Lu, Ziheng and Li, Yuqi and Bian, Jiang},
year = {2023},
issn = {23318422},
URL = {http://dx.doi.org/10.48550/arXiv.2310.14714}
}

@article{20191406719359,
language = {English},
copyright = {Compilation and indexing terms, Copyright 2024 Elsevier Inc.},
copyright = {Compendex},
title = {Data-driven prediction of battery cycle life before capacity degradation},
journal = {Nature Energy},
author = {Severson, Kristen A. and Attia, Peter M. and Jin, Norman and Perkins, Nicholas and Jiang, Benben and Yang, Zi and Chen, Michael H. and Aykol, Muratahan and Herring, Patrick K. and Fraggedakis, Dimitrios and Bazant, Martin Z. and Harris, Stephen J. and Chueh, William C. and Braatz, Richard D.},
volume = {4},
number = {5},
year = {2019},
pages = {383 - 391},
issn = {20587546}
}

@article{20221311845974 ,
language = {English},
copyright = {Compilation and indexing terms, Copyright 2024 Elsevier Inc.},
copyright = {Compendex},
title = {Sensing as the key to battery lifetime and sustainability},
journal = {Nature Sustainability},
author = {Huang, Jiaqiang and Boles, Steven T. and Tarascon, Jean-Marie},
volume = {5},
number = {3},
year = {2022},
pages = {194 - 204},
issn = {23989629},
URL = {http://dx.doi.org/10.1038/s41893-022-00859-y}
}

@inproceedings{Warpformer,
author = {Zhang, Jiawen and Zheng, Shun and Cao, Wei and Bian, Jiang and Li, Jia},
title = {Warpformer: A Multi-scale Modeling Approach for Irregular Clinical Time Series},
year = {2023},
isbn = {9798400701030},
publisher = {Association for Computing Machinery},
address = {New York, NY, USA},
url = {https://doi.org/10.1145/3580305.3599543},
doi = {10.1145/3580305.3599543},
booktitle = {Proceedings of the 29th ACM SIGKDD Conference on Knowledge Discovery and Data Mining},
pages = {3273–3285},
numpages = {13},
location = {Long Beach, CA, USA},
series = {KDD '23}
}

@article{Informer, 
title={Informer: Beyond Efficient Transformer for Long Sequence Time-Series Forecasting}, volume={35}, 
url={https://ojs.aaai.org/index.php/AAAI/article/view/17325}, 
DOI={10.1609/aaai.v35i12.17325}, 
number={12}, 
journal={Proceedings of the AAAI Conference on Artificial Intelligence}, author={Zhou, Haoyi and Zhang, Shanghang and Peng, Jieqi and Zhang, Shuai and Li, Jianxin and Xiong, Hui and Zhang, Wancai}, year={2021}, month={May}, pages={11106-11115} }

@Article{20182105222726,
AUTHOR = {Devie, Arnaud and Baure, George and Dubarry, Matthieu},
TITLE = {Intrinsic Variability in the Degradation of a Batch of Commercial 18650 Lithium-Ion Cells},
JOURNAL = {Energies},
VOLUME = {11},
YEAR = {2018},
NUMBER = {5},
ARTICLE-NUMBER = {1031},
URL = {https://www.mdpi.com/1996-1073/11/5/1031},
ISSN = {1996-1073},
DOI = {10.3390/en11051031}
}

@article{20242016083489,
language = {English},
copyright = {Compilation and indexing terms, Copyright 2026 Elsevier Inc.},
copyright = {Compendex},
title = {Attention towards chemistry agnostic and explainable battery lifetime prediction},
journal = {npj Computational Materials},
author = {Rahmanian, Fuzhan and Lee, Robert M. and Linzner, Dominik and Michel, Kathrin and Merker, Leon and Berkes, Balazs B. and Nuss, Leah and Stein, Helge Soren},
volume = {10},
number = {1},
year = {2024},
issn = {20573960},
URL = {http://dx.doi.org/10.1038/s41524-024-01286-7}
}

@article{20250417745793,
language = {English},
copyright = {Compilation and indexing terms, Copyright 2026 Elsevier Inc.},
copyright = {Compendex},
title = {Non-destructive degradation pattern decoupling for early battery trajectory prediction via physics-informed learning},
journal = {Energy and Environmental Science},
author = {Tao, Shengyu and Zhang, Mengtian and Zhao, Zixi and Li, Haoyang and Ma, Ruifei and Che, Yunhong and Sun, Xin and Su, Lin and Sun, Chongbo and Chen, Xiangyu and Chang, Heng and Zhou, Shiji and Li, Zepeng and Lin, Hanyang and Liu, Yaojun and Yu, Wenjun and Xu, Zhongling and Hao, Han and Moura, Scott and Zhang, Xuan and Li, Yang and Hu, Xiaosong and Zhou, Guangmin},
volume = {18},
number = {3},
year = {2025},
pages = {1544 - 1559},
issn = {17545692},
URL = {http://dx.doi.org/10.1039/d4ee03839h},
}

@article{LI2022453,
title = {Forecasting battery capacity and power degradation with multi-task learning},
journal = {Energy Storage Materials},
volume = {53},
pages = {453-466},
year = {2022},
issn = {2405-8297},
doi = {https://doi.org/10.1016/j.ensm.2022.09.013},
url = {https://www.sciencedirect.com/science/article/pii/S2405829722004998},
author = {Weihan Li and Haotian Zhang and Bruis {van Vlijmen} and Philipp Dechent and Dirk Uwe Sauer}
}

@article{hendrycks2016gaussian,
  title={Gaussian error linear units (gelus)},
  author={Hendrycks, Dan and Gimpel, Kevin},
  journal={arXiv preprint arXiv:1606.08415},
  year={2016}
}

@INPROCEEDINGS{10378627,
  author={Tan, Shuai and Ji, Bin and Pan, Ye},
  booktitle={2023 IEEE/CVF International Conference on Computer Vision (ICCV)}, 
  title={EMMN: Emotional Motion Memory Network for Audio-driven Emotional Talking Face Generation}, 
  year={2023},
  volume={},
  number={},
  pages={22089-22099},
  doi={10.1109/ICCV51070.2023.02024}}

@misc{weston2015memorynetworks,
      title={Memory Networks}, 
      author={Jason Weston and Sumit Chopra and Antoine Bordes},
      year={2015},
      eprint={1410.3916},
      archivePrefix={arXiv},
      primaryClass={cs.AI},
      url={https://arxiv.org/abs/1410.3916}, 
}

@misc{MedSpaformer,
      title={MedSpaformer: a Transferable Transformer with Multi-granularity Token Sparsification for Medical Time Series Classification}, 
      author={Jiexia Ye and Weiqi Zhang and Ziyue Li and Jia Li and Fugee Tsung},
      year={2025},
      eprint={2503.15578},
      archivePrefix={arXiv},
      primaryClass={cs.LG},
      url={https://arxiv.org/abs/2503.15578}, 
}

@article{20254519447317 ,
language = {English},
copyright = {Compilation and indexing terms, Copyright 2026 Elsevier Inc.},
copyright = {Compendex},
title = {A Lightweight Multiscale Signal Learning Framework for Predicting Battery Degradation Trajectory},
journal = {IEEE Sensors Journal},
author = {Shen, Quanyong and Li, Jian and Nie, Jiahao and Bao, Zhengyi and Wang, Chenhan},
volume = {25},
number = {24},
year = {2025},
pages = {44801 - 44812},
issn = {1530437X},
URL = {http://dx.doi.org/10.1109/JSEN.2025.3625630}
}

@ARTICLE{10332202,
  author={Li, Jinwen and Deng, Zhongwei and Che, Yunhong and Xie, Yi and Hu, Xiaosong and Teodorescu, Remus},
  journal={IEEE Transactions on Transportation Electrification}, 
  title={Degradation Pattern Recognition and Features Extrapolation for Battery Capacity Trajectory Prediction}, 
  year={2024},
  volume={10},
  number={3},
  pages={7565-7579},
  doi={10.1109/TTE.2023.3336618}}

@article{wang2024physics,
  title={Physics-informed neural network for lithium-ion battery degradation stable modeling and prognosis},
  author={Wang, Fujin and Zhai, Zhi and Zhao, Zhibin and Di, Yi and Chen, Xuefeng},
  journal={Nature Communications},
  volume={15},
  number={1},
  pages={4332},
  year={2024},
  publisher={Nature Publishing Group UK London},
  url={https://doi.org/10.1038/s41467-024-48779-z},
}

@article{20232714333821,
language = {English},
copyright = {Compilation and indexing terms, Copyright 2026 Elsevier Inc.},
copyright = {Compendex},
title = {Lifetime Prediction of Lithium Ion Batteries by Using the Heterogeneity of Graphite Anodes},
journal = {ACS Energy Letters},
author = {Kim, Minsoo and Kim, Inwoo and Kim, Jisub and Choi, Jang Wook},
volume = {8},
number = {7},
year = {2023},
pages = {2946 - 2953},
issn = {23808195},
URL = {http://dx.doi.org/10.1021/acsenergylett.3c00695}
}

@article{20244517308214,
language = {English},
copyright = {Compilation and indexing terms, Copyright 2025 Elsevier Inc.},
copyright = {Compendex},
title = {Data-driven analysis of battery formation reveals the role of electrode utilization in extending cycle life},
journal = {Joule},
author = {Cui, Xiao and Kang, Stephen Dongmin and Wang, Sunny and Rose, Justin A. and Lian, Huada and Geslin, Alexis and Torrisi, Steven B. and Bazant, Martin Z. and Sun, Shijing and Chueh, William C.},
volume = {8},
number = {11},
year = {2024},
pages = {3072 - 3087},
issn = {25424351},
URL = {http://dx.doi.org/10.1016/j.joule.2024.07.024},
}

@article{20214611166935,
language = {English},
copyright = {Compilation and indexing terms, Copyright 2025 Elsevier Inc.},
copyright = {Compendex},
title = {Predicting the impact of formation protocols on battery lifetime immediately after manufacturing},
journal = {Joule},
author = {Weng, Andrew and Mohtat, Peyman and Attia, Peter M. and Sulzer, Valentin and Lee, Suhak and Less, Greg and Stefanopoulou, Anna},
volume = {5},
number = {11},
year = {2021},
pages = {2971 - 2992},
issn = {25424351},
URL = {http://dx.doi.org/10.1016/j.joule.2021.09.015},
}

@article{mohtat2021reversible,
  title={Reversible and irreversible expansion of lithium-ion batteries under a wide range of stress factors},
  author={Mohtat, Peyman and Lee, Suhak and Siegel, Jason B and Stefanopoulou, Anna G},
  journal={Journal of The Electrochemical Society},
  volume={168},
  number={10},
  pages={100520},
  year={2021},
  publisher={IOP Publishing}
}

@inproceedings{BatteryLife,
author = {Tan, Ruifeng and Hong, Weixiang and Tang, Jiayue and Lu, Xibin and Ma, Ruijun and Zheng, Xiang and Li, Jia and Huang, Jiaqiang and Zhang, Tong-Yi},
title = {BatteryLife: A Comprehensive Dataset and Benchmark for Battery Life Prediction},
year = {2025},
isbn = {9798400714542},
publisher = {Association for Computing Machinery},
address = {New York, NY, USA},
url = {https://doi.org/10.1145/3711896.3737372},
doi = {10.1145/3711896.3737372},
booktitle = {Proceedings of the 31st ACM SIGKDD Conference on Knowledge Discovery and Data Mining V.2},
pages = {5789–5800},
numpages = {12},
location = {Toronto ON, Canada},
series = {KDD '25}
}

@article{20203809192263,
doi = {10.1149/1945-7111/abae37},
url = {https://dx.doi.org/10.1149/1945-7111/abae37},
year = {2020},
month = {sep},
publisher = {IOP Publishing},
volume = {167},
number = {12},
pages = {120532},
author = {Preger, Yuliya and Barkholtz, Heather M. and Fresquez, Armando and Campbell, Daniel L. and Juba, Benjamin W. and Romàn-Kustas, Jessica and Ferreira, Summer R. and Chalamala, Babu},
title = {Degradation of Commercial Lithium-Ion Cells as a Function of Chemistry and Cycling Conditions},
journal = {Journal of The Electrochemical Society}
}

@online{batteryarchive,
  year  =        2024,
  title =        "BatteryArchive.org.",
  url =          "https://batteryarchive.org/index.html",
}

@article{liu2025timebridge,
      title={TimeBridge: Non-Stationarity Matters for Long-term Time Series Forecasting}, 
      author={Liu, Peiyuan and Wu, Beiliang and Hu, Yifan and Li, Naiqi and Dai, Tao and Bao, Jigang and Xia, Shu-Tao},
      journal={International Conference on Machine Learning},
      year={2025},
}

@article{ZHENG2026114180,
title = {Self-discharge estimation for lithium-ion batteries based on formation data in production},
journal = {Engineering Applications of Artificial Intelligence},
volume = {169},
pages = {114180},
year = {2026},
issn = {0952-1976},
doi = {https://doi.org/10.1016/j.engappai.2026.114180},
url = {https://www.sciencedirect.com/science/article/pii/S0952197626004616},
author = {Haoyuan Zheng and Shaobin Yang and Weihua Xue and Shouzhen Xiao and Ding Shen and Wei Dong and Xu Zhang}
}

\appendix
\section{End-of-life Definition}\label{AppendixEOL}
We define end-of-life $t_{\mathrm{eol}}$ as the first cycle at which $\mathrm{SOH}$ falls below the threshold $\tau$.
We use $\tau=80\%$ for Li-ion, Na-ion, and Zn-ion.
For CALB, since many batteries are not degraded to $80\%$ SOH within the measured data, we use $\tau=90\%$ following BatteryLife \cite{BatteryLife}.

\section{Details of SOC Calculation}\label{AppendixSOC}
Each battery provides a protocol SOC interval $\bigl[\mathrm{SOC}^{\mathrm{ch}}_{\mathrm{start}},\,\mathrm{SOC}^{\mathrm{ch}}_{\mathrm{end}}\bigr]$, where charging starts at $\mathrm{SOC}^{\mathrm{ch}}_{\mathrm{start}}$ and ends at $\mathrm{SOC}^{\mathrm{ch}}_{\mathrm{end}}$; the subsequent discharge returns to $\mathrm{SOC}^{\mathrm{ch}}_{\mathrm{start}}$ before the next cycle.

Within each cycle $i$, we assume SOC varies linearly with the within-segment charge/discharge capacity change.
Let $Q_{i,k}$ denote the capacity at point $k$ in cycle $i$.
For charging, with segment endpoints $Q^{\mathrm{ch}}_{i,\mathrm{start}}$ and $Q^{\mathrm{ch}}_{i,\mathrm{end}}$, we compute
\begin{equation}
\mathrm{SOC}_{i,k}
=
\mathrm{SOC}^{\mathrm{ch}}_{\mathrm{start}}
+
\frac{Q_{i,k}-Q^{\mathrm{ch}}_{i,\mathrm{start}}}{Q^{\mathrm{ch}}_{i,\mathrm{end}}-Q^{\mathrm{ch}}_{i,\mathrm{start}}}
\left(\mathrm{SOC}^{\mathrm{ch}}_{\mathrm{end}}-\mathrm{SOC}^{\mathrm{ch}}_{\mathrm{start}}\right).
\label{eq:soc_charge}
\end{equation}
For discharging, we use the same linear mapping but reverse the SOC direction from $\mathrm{SOC}^{\mathrm{ch}}_{\mathrm{end}}$ to $\mathrm{SOC}^{\mathrm{ch}}_{\mathrm{start}}$:
\begin{equation}
\mathrm{SOC}_{i,k}
=
\mathrm{SOC}^{\mathrm{ch}}_{\mathrm{end}}
+
\frac{Q_{i,k}-Q^{\mathrm{dis}}_{i,\mathrm{start}}}{Q^{\mathrm{dis}}_{i,\mathrm{end}}-Q^{\mathrm{dis}}_{i,\mathrm{start}}}
\left(\mathrm{SOC}^{\mathrm{ch}}_{\mathrm{start}}-\mathrm{SOC}^{\mathrm{ch}}_{\mathrm{end}}\right).
\label{eq:soc_discharge}
\end{equation}

\textbf{SOC-aligned resampling.}
We re-parameterize each segment by SOC and resample voltage, current, and capacity on a uniform SOC grid.
Concretely, for charging we interpolate each variable at $L/2$ equally spaced SOC values in $\bigl[\mathrm{SOC}^{\mathrm{ch}}_{\mathrm{start}},\,\mathrm{SOC}^{\mathrm{ch}}_{\mathrm{end}}\bigr]$; for discharging we use $L/2$ equally spaced values in the reverse direction.
We then concatenate the resampled charging and discharging sequences to form a length-$L$ per-cycle input, where the $k$-th point corresponds to a fixed SOC level (within the recorded interval), yielding SOC-aligned inputs for BatteryMFormer.

\section{Further Implementation Details}\label{Appendix:Implementation_details}
This appendix provides additional implementation details to facilitate reproducibility.
We first describe the unified input processing pipeline used for all baselines that leverage cycling data, and then explain how we adapt generic time-series forecasters and battery-specific baselines to the early BDTF setting.

\subsection{Input and Target Construction}
\textbf{Input processing}.
For each battery, we process the raw cycling record on a per-cycle basis.
For cycle $i$, we resample the charging and discharging segments to $L/2$ uniformly spaced points (per segment) and concatenate them in the order \emph{charge $\rightarrow$ discharge}, yielding length-$L$ sequences for voltage, current, and capacity.
Let $\mathbf{v}_i\in\mathbb{R}^{L}$, $\mathbf{I}_i\in\mathbb{R}^{L}$, and $\mathbf{c}_i\in\mathbb{R}^{L}$ denote the resampled voltage, current, and capacity, respectively.
Following prior work \cite{BatteryLife}, we normalize current to C-rate by dividing by the (per-battery) nominal capacity:
\begin{equation}
\mathbf{I}_i \leftarrow \mathbf{I}_i \,/\, Q_{\mathrm{nominal}},
\end{equation}
where $Q_{\mathrm{nominal}}$ is provided by the dataset.
Voltage and capacity are kept in their original scales.
We then form the per-cycle input as
\begin{equation}
\bar{\mathbf{X}}_i = [\mathbf{v}_i;\,\mathbf{I}_i;\,\mathbf{c}_i]\in\mathbb{R}^{3\times L}.
\end{equation}

For BatteryMFormer, we additionally compute the SOC variable as described in Appendix~\ref{AppendixSOC}.
All models are trained and evaluated under the early BDTF protocol in Section~\ref{SecTaskFormulation}:
we use at most the first 100 cycles as input, and if fewer cycles are available, we pad the missing cycles with all-zero sequences.
Specifically, for a setting with $S\in\{1,\ldots,100\}$ usable early cycles, we build $\bar{\mathbf{X}}_{1:100}$ by placing the available $\{\bar{\mathbf{X}}_i\}_{i=1}^{S}$ in the first $S$ cycles and zero-padding the remaining cycles.
We also provide a cycle-level validity mask
\begin{equation}
\mathbf{m}^{\mathrm{cyc}}\in\{0,1\}^{100},
\end{equation}
where $m^{\mathrm{cyc}}_i=1$ if and only if cycle $i$ exists (i.e., $i\le S$) and $0$ otherwise; models that support attention masking use $\mathbf{m}^{\mathrm{cyc}}$ to ignore padded cycles.

\textbf{Target normalization and padding}.
Each battery trajectory is padded to a maximum horizon of 5000 cycles, which covers the longest trajectories in the database.
To account for domain-specific EOL thresholds, we normalize SOH as
\begin{equation}
\tilde{y}_j = \frac{y_j - \tau}{1-\tau},
\end{equation}
where $y_j$ is the ground-truth SOH at cycle $j$ and $\tau$ is the EOL threshold defined in Appendix~\ref{AppendixEOL}. During training, Equation (\ref{loss:pred}) is applied in the normalized SOH space using $\tilde{\mathbf{y}}$ and $\hat{\tilde{\mathbf{y}}}$. For evaluation, predictions are transformed back to the original SOH scale before computing MAE and MAPE.

\subsection{Baseline Implementation}\label{Appendix:Baseline_implementation}
We next describe how we adapt baselines for battery degradation trajectory forecasting.

\textbf{Generic time-series forecasting models.}
We reshape $\bar{\mathbf{X}}_{1:100}\in\mathbb{R}^{100\times 3\times L}$ as the time series inputs of the generic time-series forecasting models:
\begin{equation}
\mathbf{X}_{\mathrm{flat}} = \mathrm{Reshape}(\bar{\mathbf{X}}_{1:100}) \in \mathbb{R}^{(100\cdot L)\times 3}.
\end{equation}
The core architecture of the backbone encoder $f(\cdot)$ is kept unchanged from the original implementation, and we replace the original forecasting head with a trajectory prediction head that outputs a length-5000 SOH sequence:
\begin{gather}
\mathbf{H}=f(\mathbf{X}_{\mathrm{flat}}), \\
\hat{\tilde{\mathbf{y}}}=\mathrm{Head}(\mathbf{H})\in\mathbb{R}^{5000}.
\end{gather}
Here, $\mathrm{Head}(\cdot)$ first flattens $\mathbf{H}$ if necessary and then applies a linear projection to forecast 5000 SOH points.

TimesFM is a pre-trained time-series foundation model that performs zero-shot forecasting without task-specific fine-tuning. Since TimesFM only supports univariate time-series inputs, we feed the historical SOH sequence directly into the model. Due to its patch-based architecture with a fixed patch length of 32, the input length must be a multiple of 32. If the length of the historical SOH sequence is not a multiple of 32, we pad the sequence to the next multiple of 32 by repeating the last observed SOH value. TimesFM then autoregressively generates the future SOH sequence up to the specified prediction horizon.

\textbf{Battery-specific models.}
For CPTransformer and CPMLP, we follow the official BatteryLife implementations and only adjust the output head to produce the length-5000 trajectory.
For IC2ML~\cite{HUANG2026239148}, the original method constructs the input from the charging capacity-increment sequence within a fixed 3.6--3.8~V voltage window. Because this fixed voltage window is not available or comparable for all batteries and protocols in our datasets, we adapt IC2ML by computing the capacity-increment sequence over each battery's observed charging voltage range and use it as the model input. All other components follow the original paper and the official repository.

\textbf{Training objective}.
For supervised baselines, we use the same masked-MSE prediction loss as BatteryMFormer (Equation~\ref{loss:pred}). 
For IC2ML, we additionally include its original multi-task auxiliary losses. 
TimesFM is evaluated zero-shot without task-specific fine-tuning.

\textbf{Hyperparameter search}.
For BatteryMFormer, we perform per-fold hyperparameter search using Bayesian optimization, running multiple trials per fold and selecting the configuration with the lowest validation MAPE. The search space includes learning rate in $[2\times10^{-5}, 2\times10^{-4}]$, batch size in $\{64,128\}$, dropout rate in $[0.05,0.5]$, embedding dimension $d\in\{64,128,256\}$, feed-forward dimensions $d_{\mathrm{ff}}\in\{32,64,128\}$ and $d_{\mathrm{ffs}}\in\{32,64,128,256\}$, key dimension in $\{512,768\}$, memory dimension in $\{128,512\}$, $L_{\mathrm{intra}}\in\{2,4\}$, $L_{de}\in\{2,4,6,8\}$, number of queries in $\{4,8,10,12,20,50\}$, $N_{\mathrm{mem}}\in\{64,96\}$, and patch-encoder kernel size in $\{10,16,20,30\}$. For fair comparison, all models except the zero-shot TimesFM are tuned on the same training/validation splits with at least 10 configurations per domain, and the configuration with the lowest validation MAPE is selected for test evaluation.

\section{Further Details of Data Preprocessing}\label{Appendix:Data_processing}
This section describes the preprocessing pipeline applied to all datasets. 
Raw operational data may contain non-standard segments, such as reference performance tests (RPTs), formation cycles, and equipment faults. 
These segments can introduce abrupt SOH deviations that are unrelated to the predominant cycling protocol and may impair model training if treated as regular degradation signals. 
We therefore smooth evident measurement artifacts and non-standard testing segments, while preserving sustained degradation trends.

\textbf{SOH computation and battery filtering.}
We compute $\mathrm{SOH}$ as defined in Section~\ref{SecDegradation}. 
To suppress isolated measurement artifacts, single-cycle SOH drops exceeding 3\% of the previous cycle's SOH are clipped to the previous-cycle value. 
This operation is only applied to isolated spike-like drops and does not modify sustained degradation trends. 
Batteries whose SOH has not degraded below $\tau + 2.5\%$ are excluded due to insufficient degradation information, where $\tau = 90\%$ for CALB~\cite{BatteryLife} and $\tau = 80\%$ for all other datasets (Appendix~\ref{AppendixEOL}). 
For batteries that have degraded to $\tau + 2.5\%$ but have not yet reached $\tau$, we estimate the EOL point by linearly extrapolating the last 20 measured cycles to $\tau$. 
Measured SOH values are used whenever available; the extrapolated segment is only used to complete the trajectory up to the estimated EOL point for consistent target construction.

\textbf{SOH trajectory smoothing.}
We identify artifact onsets at the region level using dataset-specific information. 
For datasets with RPT timestamps (ISU-ILCC~\cite{LI2024101891}), each RPT start time is mapped to a cycle index, and the last normal cycle before the RPT is used as the onset. 
For datasets with cycle timestamps but no RPT annotations (HNEI~\cite{20182105222726}, RWTH~\cite{LI2021230024}, Tongji~\cite{zhu2022data}, MICH\_EXP~\cite{mohtat2021reversible}), cycle $k$ is flagged as an onset when $(t_k - t_{k-1}) > \gamma_{\mathrm{gap}}$, where $\gamma_{\mathrm{gap}}$ is fixed for each dataset before model training. 
For the remaining datasets, cycle $k$ is flagged when $\delta_k > \gamma^{+}$ or $\delta_k < \gamma^{-}$, where $\delta_k = (\mathrm{SOH}_k - \mathrm{SOH}_{k-1}) / \mathrm{SOH}_{k-1}$, and $\gamma^{+}$ and $\gamma^{-}$ are the 99th and 1st percentiles of the empirical $\{\delta_k\}$ distribution computed on the training split.

After an artifact onset $k_s$ is identified, we locate the recovery point $k_e$ by scanning forward and selecting the earliest cycle whose SOH returns within a tolerance $\epsilon$ of the pre-artifact level $\mathrm{SOH}_{k_s-1}$ and remains within this tolerance for the next $W$ consecutive cycles. 
The affected region $[k_s, k_e]$ is then smoothed using PCHIP (Piecewise Cubic Hermite Interpolating Polynomial) interpolation over cycle indices, with $M$ normal cycles immediately before and after the region used as anchor points. 
The same preprocessing pipeline is applied consistently across training, validation, and test splits.

\end{document}